\documentclass[journal]{IEEEtran}
\usepackage{amsmath,amsfonts}
\usepackage{algorithmic}
\usepackage{array}
\usepackage[caption=false,font=normalsize,labelfont=sf,textfont=sf]{subfig}
\usepackage{textcomp}
\usepackage{stfloats}
\usepackage{url}
\usepackage{verbatim}
\usepackage{graphicx}
\def\BibTeX{{\rm B\kern-.05em{\sc i\kern-.025em b}\kern-.08em
    T\kern-.1667em\lower.7ex\hbox{E}\kern-.125emX}}
\usepackage{balance}

\usepackage[numbers]{natbib}
\usepackage{multicol}
\usepackage{multirow} 
\usepackage{bm}
\usepackage{booktabs}
\usepackage[table]{xcolor}
\usepackage[scr=boondox,cal=esstix]{mathalpha}

\usepackage{algorithm}  
\usepackage{algorithmic}  

\usepackage{color}

\begin{document}

\title{High-fidelity 3D Gaussian Inpainting: preserving multi-view consistency and photorealistic details}

\author{
~\IEEEauthorblockN{Jun Zhou, Dinghao Li, Nannan Li, Mingjie Wang} \\
\thanks{The authors would like to thank the High Performance Computing Center of Dalian Maritime University for providing the computing resources. This work was supported by NSFC (No.~62002040) and Fundamental Research Funds for the Central Universities (No.~3132025274).(Corresponding author: Jun Zhou.)} 
\thanks{J. Zhou, D. Li, and N. Li are with the School of Information Science and Technology, Dalian Maritime University, Dalian, China (E-mail: jun90@dlmu.edu.cn, ldh123@dlmu.edu.cn, nannanli@dlmu.edu.cn). }
\thanks{M. Wang is with the School of Science, Zhejiang Sci-Tech University, Zhe Jiang, China (E-mail: mingjiew@zstu.edu.cn). }
}


\maketitle

\begin{abstract}
Recent advancements in multi-view 3D reconstruction and novel-view synthesis, particularly through Neural Radiance Fields (NeRF) and 3D Gaussian Splatting (3DGS), have greatly enhanced the fidelity and efficiency of 3D content creation. However, inpainting 3D scenes remains a challenging task due to the inherent irregularity of 3D structures and the critical need for maintaining multi-view consistency. In this work, we propose a novel 3D Gaussian inpainting framework that reconstructs complete 3D scenes by leveraging sparse inpainted views. Our framework incorporates an automatic Mask Refinement Process and region-wise Uncertainty-guided Optimization. Specifically, we refine the inpainting mask using a series of operations, including Gaussian scene filtering and back-projection, enabling more accurate localization of occluded regions and realistic boundary restoration. Furthermore, our Uncertainty-guided Fine-grained Optimization strategy, which estimates the importance of each region across multi-view images during training, alleviates multi-view inconsistencies and enhances the fidelity of fine details in the inpainted results. Comprehensive experiments conducted on diverse datasets demonstrate that our approach outperforms existing state-of-the-art methods in both visual quality and view consistency. 
\end{abstract}


\section{Introduction}
Multi-view 3D reconstruction and novel-view synthesis are crucial for creating high-fidelity 3D content of real-world scenes, enabling applications such as telepresence and AR/VR. Recent advancements in Neural Radiance Fields (NeRF)\cite{mildenhall2021nerf,barron2021mip,wang2021neus} and 3D Gaussian Splatting (3DGS)\cite{kerbl20233d,huang20242d,guedon2024sugar,lu2024scaffold} have significantly accelerated progress in this field. Among these, 3D Gaussian-based approaches have garnered substantial attention due to their ability to produce photorealistic images while achieving impressive rendering speeds. By leveraging the advantages of 3D Gaussian Splatting, researchers can more easily construct and generate richer 3D~\cite{chung2023luciddreamer,yi2023gaussiandreamer,yi2023gaussiandreamer} and even 4D assets~\cite{wu20244d,yang2023real,lin2024gaussian}, as well as reconstruct physical laws~\cite{zhang2025physdreamer,xie2024physgaussian} from scenes to enable interactive engagement~\cite{jiang2024vr}. As a result, the demand for technologies that facilitate the editing and manipulation of such scenes has surged. Among these, 3D scene inpainting has emerged as a prominent research focus. While inpainting techniques have been extensively studied in 2D image domains~\cite{jam2021comprehensive}, the challenge of inpainting 3D scenes remains significant due to the multi-view nature of scene data and the irregular structures inherent in 3D representations. These complexities make 3D scene inpainting both a demanding and promising area of exploration.

\begin{figure*} [htbp]
\centering
\includegraphics[width=0.81\textwidth]{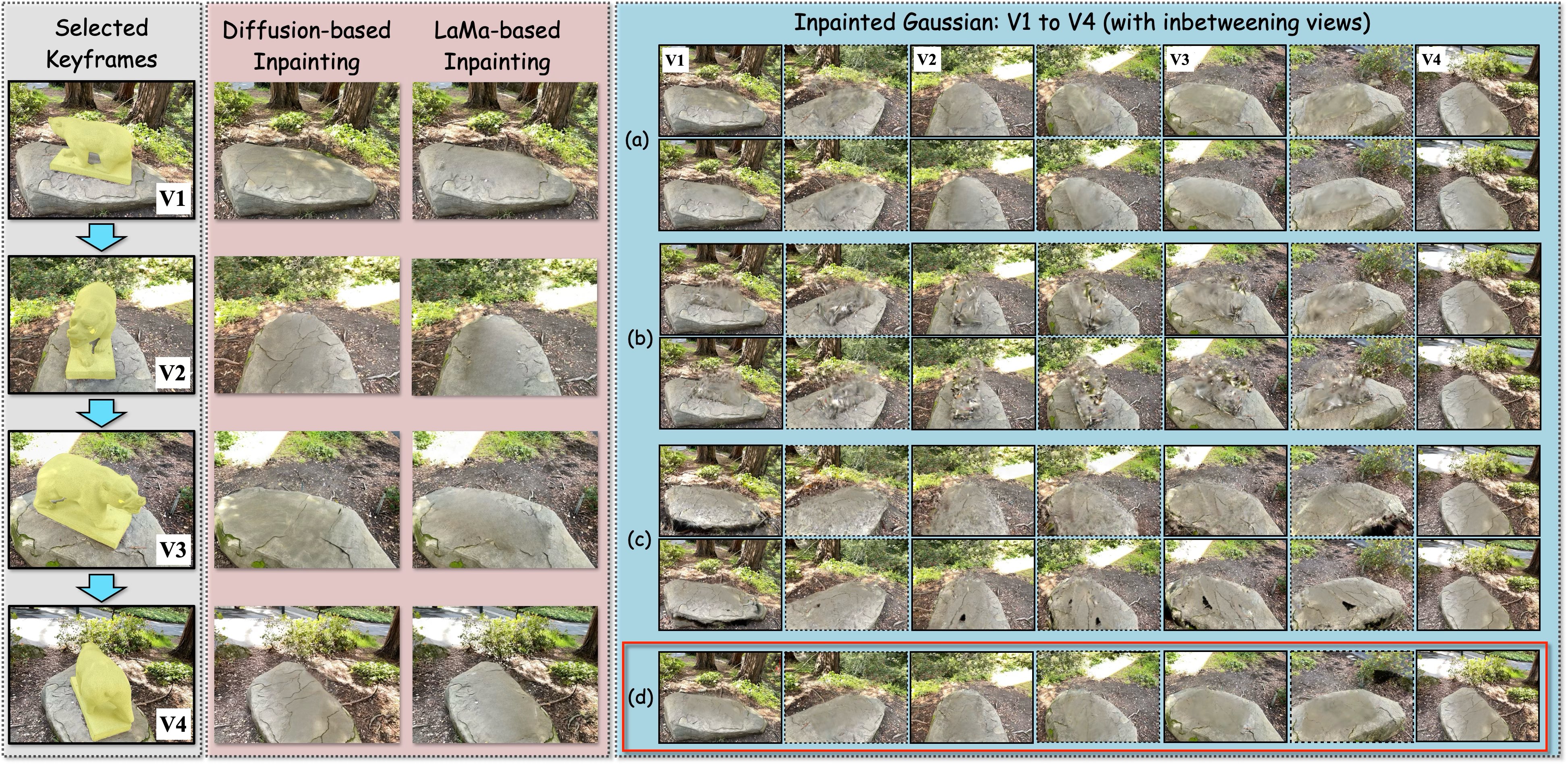}
\caption{Comparison of Different Optimization Strategies for 3D Gaussian Inpainting: (1) The left section shows the four keyframes used for training; (2) the middle section compares inpainted keyframes using LaMa~\cite{suvorov2022resolution} and a diffusion-based method~\cite{rombach2022high}; (3) the right section presents the final 3D Gaussian scene inpainting results from seven viewpoints, including keyframes and intermediate views. Specifically, (a) compares the results using all keyframes with diffusion-based (first row) and LaMa-based (second row) methods; (b) compares progressive and single-view (view V$4$) strategies; (c) showcases results from InFusion~\cite{liu2024infusion} using both progressive and single-view (view V$4$) approaches; and (d) highlights the results of our proposed method, which strikes a better balance between multi-view consistency and the preservation of realistic scene details.}
\label{fig:teaser}
\end{figure*}

As pioneering works, Remove-NeRF~\cite{weder2023removing} and SPIn-NeRF~\cite{mirzaei2023spin} have demonstrated object removal and scene inpainting using NeRF representations~\cite{mildenhall2021nerf}. These methods introduced a 2D-to-3D strategy for 3D scene inpainting, utilizing LaMa~\cite{suvorov2022resolution} for object removal and inpainting across multi-view images, followed by optimizing the 3D scene with a view-consistency constraint. However, the LaMa-based inpainting technique tends to introduce blurriness in the inpainted images and lacks fine image details, while NeRF-based optimization requires considerable time, resulting in reduced temporal efficiency.

In recent years, driven by the efficiency advantages of 3D Gaussian representations~\cite{kerbl20233d}, a variety of 3D scene editing and inpainting techniques~\cite{wang2024gaussianeditor, chen2024gaussianeditor, ye2025gaussian, huang2024point, liu2024infusion} based on Gaussian representations have been extensively explored. Among these, InFusion~\cite{liu2024infusion} stands out, leveraging a latent diffusion model~\cite{rombach2022high} to simultaneously inpaint both RGB and depth images, achieving richer image details. While the diffusion model enhances detail restoration compared to LaMa, it introduces increased multi-view inconsistencies. As illustrated in Fig.\ref{fig:teaser}, we select four sparse keyframes with distinct viewpoints for 3DGS Inpainting. The results show that LaMa-based inpainting produces blurrier images with less realistic detail compared to diffusion model. However, LaMa ensures better multi-view consistency, helping to avoid inconsistencies in 3DGS Inpainting at the cost of detail richness, which results in less realistic outcomes (Fig.\ref{fig:teaser} (a), second row on the right). In contrast, diffusion-based techniques offer richer details but introduce significant multi-view inconsistencies. When trained with multiple viewpoints, these inconsistencies are exacerbated, leading to blurry results (Fig.~\ref{fig:teaser} (a), first row on the right).

To address the issue of multi-view inconsistencies in 3DGS Inpainting, we explored two optimization strategies: single-view supervision and the InFusion method, which jointly optimizes RGB and depth images, as illustrated in Fig.~\ref{fig:teaser} (b) and (c), second row. In both cases, training was conducted solely using the image from viewpoint V$4$. While these approaches produced sharp results for the supervised view, they often led to blurriness or missing content in other views, particularly in complex, large-scale scenes. This performance degradation is primarily due to the limited information provided by single-view supervision, which causes overfitting to the supervised viewpoint. We further experimented with a progressive training strategy, shown in the first rows of (b) and (c), where additional views were introduced over time. However, whether using basic image supervision or InFusion’s progressive method, earlier view information tended to be forgotten in later stages, resulting in blurred transitional views and unsuccessful scene inpainting. To overcome these limitations, we propose a novel Uncertainty-guided Fine-grained Optimization strategy by depth-based view selection. By selectively extracting informative regions across sparse viewpoints, our method achieves a better balance between multi-view consistency and realistic detail preservation. As shown in Fig.\ref{fig:teaser} (d), our approach yields more coherent and visually convincing results. While conceptually similar to the multi-view selection strategy in Remove-NeRF\cite{weder2023removing}, our method differs in two key aspects: (1) it relies on sparse supervision from a few key views, and (2) it performs fine-grained region-wise selection based on depth, where regions closer to the camera are assigned higher confidence due to their greater reliability.

Furthermore, to enhance both the efficiency and accuracy of 3DGS Inpainting, we designed a more precise automatic Mask Refinement algorithm. Specifically, we propose a mask optimization strategy that adjusts and refines the initial segmentation mask to minimize its size while preserving occluded real scene information more accurately. This refinement provides a solid and accurate data foundation for subsequent inpainting processes, significantly improving the precision and realism of the results.

\textbf{In summary, our main contributions are as follows:} We present a novel 3DGS Inpainting framework specifically designed for sparse-view inputs. This framework integrates an automatic Mask Refinement Process that extracts additional effective background information, along with a depth-based Uncertainty-guided Fine-grained Optimization strategy that strikes a balance between multi-view consistency and the preservation of rich visual details. Extensive experiments conducted on multiple benchmark datasets demonstrate that our method outperforms existing state-of-the-art approaches in the 3DGS Inpainting task.
\label{sec:intro}
\section{Related Work}

\subsection{Image and Video Inpainting}
In the field of computer vision~\cite{quan2024deep}, video and image inpainting aims to restore missing regions while ensuring seamless blending and realistic details. Early methods~\cite{elharrouss2020image,ballester2001filling,tschumperle2005vector,efros1999texture,barnes2009patchmatch,darabi2012image,huang2014image,herling2014high,guo2017patch} relied on local background cues and optimization techniques, while traditional video inpainting approaches~\cite{wexler2007space,granados2012background,newson2014video,huang2016temporally} extended these ideas to handle temporal consistency. However, they often failed in cases with large missing areas or complex motion.
Recently, deep learning methods, especially those based on transformers and diffusion models~\cite{suvorov2022resolution,rombach2022high}, have achieved impressive results, effectively overcoming these limitations. In this work, we adopt these methods to generate multi-view inpainting results, taking advantage of its ability to produce highly realistic and visually coherent images. However, compared to conventional image or video inpainting, 3D scene inpainting presents additional challenges. One of the most critical difficulties lies in maintaining consistency across multiple views while simultaneously preserving fine-grained realism from different viewpoints.

\subsection{Radiance Fields and Rendering}
Photorealistic view synthesis is a long-standing challenge in computer vision and computer graphics. Traditional 3D representations such as meshes and point clouds remain widely used due to their explicit geometry and efficient GPU-based rasterization. In recent years, Neural Fields have emerged as a powerful alternative, offering seamless integration with deep learning frameworks and enabling high-quality novel view synthesis. Neural Fields can generally be divided into three main types. Early methods~\cite{mildenhall2021nerf,barron2021mip,xiangli2022bungeenerf} model radiance fields with MLPs for high-quality view synthesis, but suffer from slow rendering due to dense ray sampling. Acceleration techniques alleviate this but often increase memory usage or degrade visual fidelity. Then, the grid-based methods~\cite{fridovich2023k, muller2022instant} discretize space into voxel or hash grids to enable fast interpolation-based rendering, offering efficiency gains but still requiring many samples and struggling with empty space representation.  Building on 3DGS, a series of follow-up works, including 2DGS~\cite{Huang2DGS2024} and Scaffold-GS~\cite{lu2024scaffold}, have introduced further enhancements. Our work is also based on the 3DGS representation.

\subsection{Radiance Fields Inpainting}
In recent years, 3D Radiance Fields have gradually emerged as a novel 3D representation, driving an increasing demand for 3D editing~\cite{chen2024gaussianeditor,ye2025gaussian,guedon2025gaussian,xu2024tiger,zhang20243ditscene, wang2024gaussianeditor}. These techniques support a variety of scene editing operations, from object replacement to appearance adjustments, granting users improved control and flexibility in 3D scene manipulation. 3D Radiance Fields  Inpainting is one prominent application that can restore missing regions in 3D scenes, ensuring high-quality and consistent multi-view rendering results. Notably, although the aforementioned editing techniques mention inpainting in 3D scenes, they primarily treat it as a post-processing step by applying image inpainting methods to the removed regions. This approach does not perform genuine inpainting directly on the 3D scene, leaving the consistency and integrity of the reconstructed scene unaddressed. 

As pioneering efforts, NeRF-In~\cite{liu2022nerf} and SPIn-NeRF~\cite{mirzaei2023spin} utilize multi-view images to restore NeRF representations. However, they fail to address the multi-view consistency issues that arise from discrepancies in the inpainted regions. To tackle this, View-Substitute~\cite{mirzaei2023reference} proposes inpainting a single reference view and guiding the synthesis of other views via depth warping and bilateral filtering to ensure consistency. Nevertheless, its reliance on a single-view reference limits performance when dealing with complex or large missing regions. Subsequent works like Removal-NeRF~\cite{weder2023removing} enhance consistency through confidence-based view selection, while OR-NeRF~\cite{yin2023or} introduces efficient multi-view segmentation combined with an integrated TensoRF framework to achieve higher-quality rendering. Although these techniques have achieved notable progress, the emergence of 3DGS has highlighted the growing demand for faster inpainting methods leveraging point-based rendering techniques.

\begin{figure*} [t]
\centering
\includegraphics[width=0.81\textwidth]{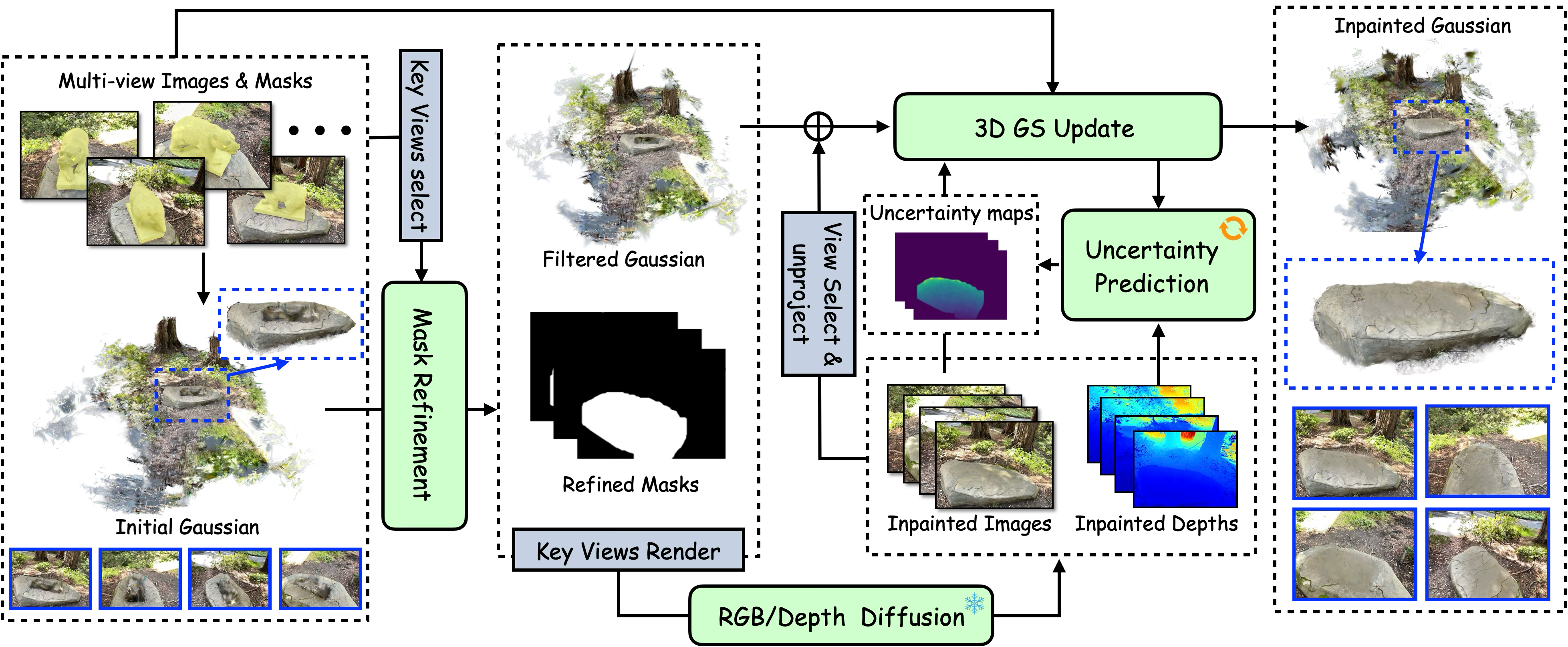}
\caption{Overview of our 3D Gaussian inpainting pipeline with Mask Refinement Process and Uncertainty-guided Optimization. Given a set of posed input images and their coarse binary masks, we first perform an initial training of the 3D Gaussian scene representation. Based on this initial representation, we introduce an automatic Mask Refinement module that accurately localizes regions requiring inpainting. In the second stage, we perform Uncertainty-guided Optimization, which selectively utilizes reliable supervision from inpainted images. This strategy effectively mitigates conflicts arising from multi-view inconsistencies and leads to a more coherent and photo-realistic 3D scene synthesis.}
\label{fig:pipeline}
\end{figure*}

Among these 3DGS inpainting methods, InFusion~\cite{liu2024infusion} employs a depth-generative diffusion model to synthesize RGB-D point clouds, which are then fused with the missing 3DGS regions for inpainting. However, this method still struggles to effectively balance high fidelity and multi-view consistency. Lu et al.\cite{lu2024view} proposed a technique similar to View-Substitute, focusing on repairing a single keyframe and using depth projection to construct consistent data for other views. Yet, it still faces difficulties in handling complex and large-scale missing regions. Wang et al.\cite{wang2025learning} used Scaffold-GS~\cite{lu2024scaffold} as the backbone and introduced an attention mechanism to learn consistent Gaussian features for the missing regions. However, multi-view inconsistencies remain unresolved. Similarly, Point’n Move~\cite{huang2023point} leveraged 2D prompt points as interactive inputs to identify missing regions and adopted a "minimize changes" optimization strategy, akin to our mask refinement approach. Despite these efforts, it still falls short of addressing multi-view inconsistencies, limiting its ability to effectively inpaint complex and large missing areas. The method by Gaussian Group~\cite{ye2025gaussian} focuses more on the semantic segmentation of objects within the Gaussian scene. Incorrect semantics can lead to inaccurate mask estimation, resulting in suboptimal inpainting. Concurrently, Huang et al.~\cite{huang20253d} propose a depth-guided multi-view strategy for consistent inpainting. Although their multi-view mask warping yields finer masks, it overlooks rendering instability near mask edges, leading to overly smooth results. Our work tackles these challenges by emphasizing multi-view consistency and the accuracy of mask estimation, ultimately improving the quality of 3DGS inpainting results.

\label{sec:related}
\section{Method}

\subsection{Overview}
Our work aims to reconstruct and inpaint 3D scenes from multi-view images and masks, achieving consistent and photorealistic representations. Built upon 3D Gaussian Splatting~\cite{kerbl20233d}, we extend its capabilities to address 3D scene inpainting. Similar to methods like InFusion~\cite{liu2024infusion} and Gaussian Group~\cite{ye2025gaussian}, our approach adopts a two-stage pipeline. In the first stage, we reconstruct a 3D scene with "holes" by utilizing masks and multi-view inputs, recovering regions outside the missing areas by leveraging background information from other views. A brief overview of this step is provided in the preliminary section (Sec.\ref{sec:background}). In the second stage, we inpaint the missing content within the "holes" using image inpainting techniques. To improve this process, we introduce an automatic Mask Refinement method (Sec.\ref{sec:automask}) for more accurate hole definition and propose a novel training framework (Sec.\ref{sec:uncertainty_inpainting}) that incorporates depth-based uncertainty scores to balance multi-view consistency and fine-detail preservation. An overview of our full pipeline is illustrated in Fig.\ref{fig:pipeline}.

\subsection{Preliminaries: 3D Gaussian Scene Initialization with Masks}\label{sec:background}
\begin{figure} [t]
\centering
\includegraphics[width=0.83\columnwidth]{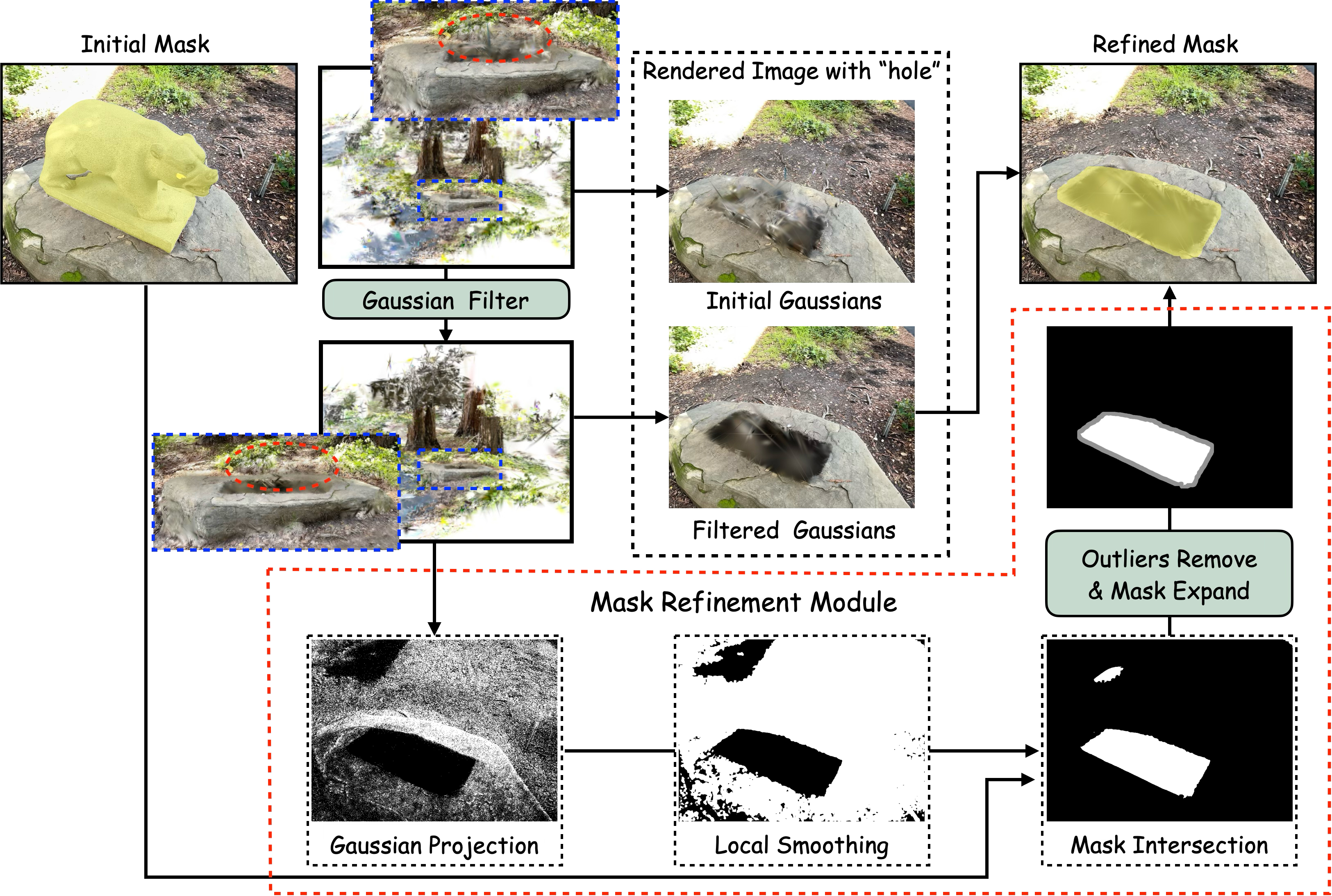}
\caption{Visualization of the Mask Refinement Process.}
\label{fig:mask_refine}
\end{figure}

The Gaussian Splitting~\cite{kerbl20233d} can be used to reconstruct 3D Gaussian representation from multi-view images. The Gaussian representation inherently possess rich geometric attributes, and can also be employed for rendering new view synthesis. Similar to Infusion~\cite{liu2024infusion}, we need to utilize both multi-view images $C^{o} = \{c^{o}_{i}\}_{i=1}^{N}$, accompanied by respective camera poses $\Pi = \{\pi_{i}\}_{i=1}^{N}$ and their corresponding masks $M = \{m_{i}\}_{i=1}^{N}$ for scene reconstruction, with the requirement to remove the masked regions. Specifically, our objective is to train an initial 3D Gaussian representation $\Theta = \{g_{i}\}_{i=1}^{L}$ with ``hole'',  and each 3D Gaussian $g_{i} $ is defined as a series attributes $g_{i} = \{ \mu_{i},s_{i},q_{i},sh_{i},\alpha_{i}\}$. Then the covariance matrix $\Sigma_{i} \in \mathbb{R}^{3 \times3}$ of the 3D Gaussian is expressed as: $\Sigma_{i}= R_{i}s_{i}s_{i}^{T}R_{i}^{T}$, where $R_{i} \in \mathbb{R}^{3 \times 3}$ is the orthogonal rotation matrix of the Gaussian parameterized by the quaternion $q_{i}$ and $s_{i} \in \mathbb{R}^{3}$ is  a scaling vector of  $g_{i}$. Once the Gaussian representation is constructed, we can project the 3D Gaussian points onto the image plane based on the given camera pose $\pi_{i} \in \Pi $. Each Gaussian $g_{j} \in \Theta $ in the collection is projected onto the image plane corresponding to the viewpoint as:
\begin{equation}
u_{j,i}^{2D} = P_{i}W_{i} \mu_{j},   \Sigma_{j,i}^{2D}= J_{j}W_{i}\Sigma_{j}^{T}W_{i}^{T}J_{j}^{T},
\end{equation}
where $\mu_{j,i}^{2D}$ and $\Sigma_{j,i}^{2D}$ respectively represent the center and the covariance matrix of the projected Gaussian distribution, $W_{i}$ represents the viewing transformation matrix, and $P_{i}$ represents the projective transformation matrix. Both can be derived from the camera pose. $J_{j}$ represents the Jacobian of the affine approximation of the projective transformation.  After that, to perform image rendering on the image plane, for each pixel $p$ of the render image $c_{i}^{r}$, its color $c_{i}^{r}(p)$ is derived through an $\alpha$-blending function as:
\begin{equation}
c_{i}^{r}(p) = \sum_{j=1}^{l} sh_{j}\beta_{j} \prod_{k=1}^{j-1}(1-\beta_{k}),
\end{equation}
where $\beta_{j} = \alpha_{j}e^{-\frac{1}{2}(p-\mu_{j,i}^{2D})^{T}(\Sigma_{j,i}^{2D})^{-1}(x-\mu_{j,i}^{2D})}$ and $l$ represents the number of projected Gaussians that overlap p. Since we only need to reconstruct the background of the 3D Gaussian scene, we apply a mask to ensure that the training process focuses exclusively on the background regions while ignoring the removed object areas. The specific loss function is formulated as follows:
\begin{equation}
\mathcal{L}_{init} = \sum_{i=1}^{N} \| c^{r}_{i}\odot m_{i}-c^{o}_{i}\odot m_{i}\|^{2},
\end{equation}
where $\odot$ is ‌Hadamard Product. Finally, after 30,000 iterations, we can obtain a 3D Gaussian representation with ``hole''. As shown in the middle part of Fig.~\ref{fig:mask_refine}, an example of rendering with missing regions (``holes'') is provided.

\subsection{Automatic Mask Refinement Process}\label{sec:automask}
As is well known, mainstream 3D inpainting methods rely on multi-view 2D image inpainting. A key step is to design an automated missing region detection algorithm for rendered images with "holes" in the first stage. A reliable algorithm should retain clear background areas while accurately masking missing regions, as shown in the final results of our method on the right side of Fig.~\ref{fig:mask_refine}. 
To address the disordered floater Gaussian kernels near the holes in the initial Gaussian (Fig.~\ref{fig:mask_refine}) that hinder accurate ``hole'' detection in rendered images, we first propose a fast filtering algorithm. Building on this, we designed a precise automatic mask refinement module, which will be detailed in the following part.

\begin{figure*} [htbp]
\centering
\includegraphics[width=0.81\textwidth]{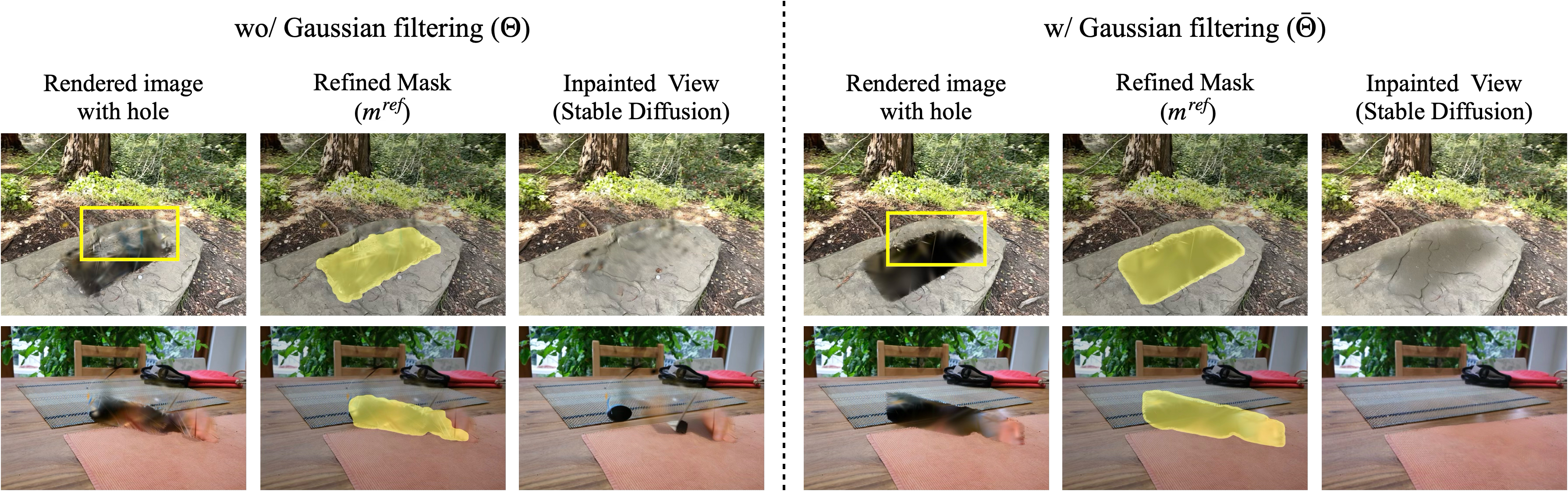}
\caption{Visualizing the effect of the Gaussian Filter: We compare the differences between the original Gaussian representation \( \Theta \) and the Gaussian representation after the Gaussian filtering operation \( \bar{\Theta} \) in terms of rendered images, the refined masks, and the inpainted images. Here, the yellow box in the figure highlights that our method effectively removes the floating Gaussians and achieves more accurate masks and reliable inpainted images.}
\label{fig:w_gs_filter}
\end{figure*}

\noindent\textbf{Gaussians Filtering.} We observed that some relatively large floating Gaussians exist in the initial Gaussian representation \(\Theta\) due to the large scope of the initial mask, as shown on the left part of the Fig.~\ref{fig:w_gs_filter}.  In other words, these floating Gaussians occur because of insufficient multi-view training outside the masks and the lack of guiding information within the missing regions. These floating Gaussians can obscure background information in certain views, thereby affecting the results of subsequent 2D inpainting, as shown in Fig.~4. To ensure a reliable 2D inpainting process, we need to remove these floating points. Thus, we assume that a valid 3D Gaussian point should never intrude into the mask area across multiple viewpoints, while a floating Gaussian point may appear inside the mask in some views. Based on this assumption, we designed a fast post-processing algorithm to remove floating kernels. .  And, this is a post-processing process that can directly accept input from a Gaussian scene. This algorithm can directly operate on our  initial 3D Gaussian scene. Specifically, given the initial Gaussian scene \(\Theta\), we select \(K\) key views \(\Pi_{K} = \{\pi_{j}\}_{j=1}^{K}\) from the set of all views \(\Pi\) to evaluate each Gaussian kernel in \(\Theta\).  For each Gaussian point \(g_{k} = \{ \mu_{k}, s_{k}, q_{k}, sh_{k}, \alpha_{k}\} \in \Theta\), its projection positions across the \(K\) key views are denoted as \(\mu^{2d}_{k,j}\). We determine whether it consistently lies outside the corresponding masks \(M_{K} = \{m_{j}\}_{j=1}^{K}\). The filter can be expressed as:
\begin{equation}
f_{\text{mask}}(g_{k})  = \prod_{j=1}^{K} m_{j}(\mu_{k,j}^{2D}),
\end{equation}
where 
\begin{equation}
m_{j}(x)  = 
\begin{cases} 
1, & \text{if } x \text{ lies outside the mask region,} \\
0, & \text{otherwise.}
\end{cases}
\end{equation}
Finally, we filter out Gaussians where \(f_{\text{mask}}(\cdot) = 0\) and obtain a new Gaussian representation \(\bar{\Theta}\). The detailed steps are presented in Algorithm~\ref{alg:algorithm_name}.

\noindent\textbf{Mask Refinement.} Since our method requires sparse-view inpainting images for 3D inpainting, a refined mask that adequately preserves the background is essential. Predicting masks  directly from rendered images with missing parts (the hole) typically relies on large models, which demand substantial computational resources. Instead, we propose constructing a reasonable mask directly from the previous filtered Gaussian representation $\bar{\Theta}$. As shown in Fig.~\ref{fig:mask_refine}, our refinement module consists of four operations: Gaussian Projection, Local Smoothing, Mask Intersection, and Mask Expansion. 

Firstly, our Gaussian Projection can projects valid Gaussian representation $\bar{\Theta}$ onto the specified viewpoints, while the non-projected regions are highly likely to contain the real mask regions. Specifically, given any camera position $\pi_{j} \in \Pi$ , each Gaussian point $\bar{g}_{k'} \in \bar{\Theta}$ 
can be projected onto the 2D space as $\mu_{k',j}^{2D}\in \mathbb{R}^{2}$. Thus, we can construct a projected image $m^{p}_{j} \in \mathbb{R}^{H\times W} $as follows:
\begin{equation}\label{eq:6}
m^{p}_{j}(x,y)  =
\begin{cases} 
1, & \text{if } \exists \, u_{k',j} ^{2D}\in [x - \frac{\epsilon}{2}, x + \ \frac{\epsilon}{2},] \times  [y- \frac{\epsilon}{2}, y+ \frac{\epsilon}{2}]  \\
0, & \text{otherwise.}
\end{cases}
\end{equation}
Here, \( (x, y) \) represents the position of any pixel, and \( \epsilon \) represents the size of a single pixel.

Secondly, a Local Smoothing operation is applied to the discrete pixels to create continuous mask regions. We perform convolution using \( 3 \times 3 \) and \( 9 \times 9 \) kernels (\( Conv_3 \) and \( Conv_9 \)) with all ones to compute the average value of pixels within a local neighborhood. Here, we apply convolution operations using smaller kernels first, followed by larger kernels, to ensure the preservation of local mask details. The smoothed projected image \( M^{s}_{j} \) is obtained by:
\begin{equation}\label{eq:7}
m^{s}_{j} = (m^{p}_{j} \ast Conv_3) \ast Conv_9.
\end{equation}

Subsequently, it is made to intersect with the initial mask $m_{j}^{inter}$. This intersection operation, denoted as $m_{j}^{inter}= (\overline{m_j^{s}} \cap m_j)$, is specifically engineered to expunge the spurious hole regions that lie outside the purview of $m_j$. In this step, our negation operation mainly sets the areas inside the mask to 0 and the areas outside the mask to 1, consistent with the initial mask. Then, we select the largest contiguous region by area as the mask region $\bar{m}_{j}^{inter} $to remove outlier areas.  

Finally, contingent upon the idiosyncratic requirements of the specific scene under consideration, an expansion operation may be deemed necessary for $\bar{m}_j^{inter}$. By designating the expansion magnitude as $\gamma$, we arrive at the ultimate refined mask, $m_j^{ref}$, which is expressed as $m_j^{ref} = Expand(\bar{m}_j^{inter}, \gamma)$.  Here $\gamma$ is set to 15.  This operation comes from a method in the OpenCV library. The complete algorithm can be found in the second stage of Algorithm~\ref{alg:algorithm_name}.

\begin{algorithm}
\caption{Automatic Mask Refinement}
\label{alg:algorithm_name}
\begin{algorithmic}[1]
	\STATE $M= \{m_{i}\}_{i=1}^{N} \gets$ SAM-Track($C= \{c_{i}\}_{i=1}^{N}$)
	\STATE $\Theta \gets$ Mask-Training($C, M$)
	\STATE $\Pi_K = \{\pi_{j}\}_{j=1}^{K} \gets$ ViewSelector($\Pi = \{\pi_{i}\}_{i=1}^{N}$)

	\STATE \textbf{\textit{Stage 1: Gaussians Filtering}}
	\FOR{$g_k = \{\mu_k, s_k, q_k, c_k, \alpha_k\}$ in $\Theta$}
		\STATE $f_{\text{mask}}(g_{k}) \gets 1$
		\FOR{$\pi_j$ in $\Pi_K$}
		\STATE $\mu^{2d}_{k,j} \gets$ proj($\mu_k, \pi_j$)
		\STATE $f_{\text{mask}}(g_{k}) = f_{\text{mask}}(g_{k})\cdot  m_j(\mu^{2d}_{k,j}) $
		\ENDFOR
	\ENDFOR
	\STATE $\bar{\Theta} \gets$ Remove($g_k$, where $f_{\text{mask}}(g_k) = 1$)
	
	\STATE  \textbf{\textit{Stage 2: Mask Refinement}}
	\FOR{$\pi_j$ in $\Pi_K$}
	\FOR{$g_k = \{\mu_k, s_k, q_k, c_k, \alpha_k\}$ in $\bar{\Theta}$}
	\STATE $\mu^{2d}_{k,j} \gets$ proj($\mu_k, \pi_j$)
	\STATE $(x,y) \gets$ Convert $\mu^{2d}_{k,j}$ to pixel coordinates
	\STATE $m_j^p \gets$ Calculate using Eq.~\ref{eq:6}
	\ENDFOR
	\STATE $m_j^s \gets$ Calculate using Eq.~\ref{eq:7}
	\STATE $m_{j}^{inter} \gets$ Intersection($m_j^s$,$m_{j}$)

	\STATE $m_j^{\text{ref}} \gets$ Expand($m_{j}^{inter}$, $\gamma$) 
	\ENDFOR
\end{algorithmic}
\end{algorithm}

\begin{figure} [t]
\centering
\includegraphics[width=0.83\columnwidth]{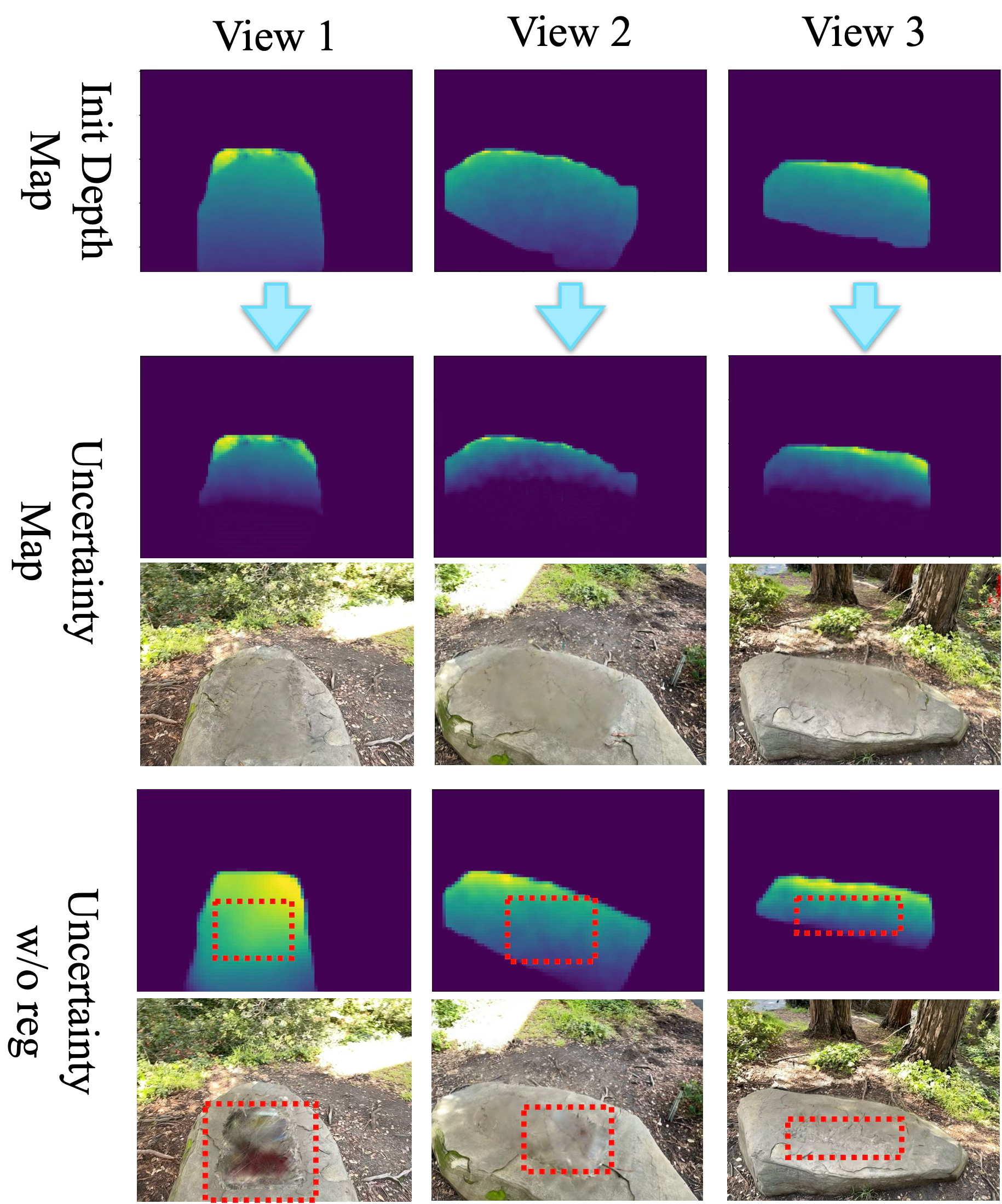}
\caption{Visualization of Uncertainty Optimization Results Initialized from Depth. We compare results with and without the regularization term. The red boxes highlight that without the  regularization term, dense uncertainty regions lead to more chaotic Gaussian field estimation.}
\label{fig:uncertainty_map}
\end{figure}

\subsection{Uncertainty-Based Sparse View Consistency Inpainting}\label{sec:uncertainty_inpainting}

\noindent\textbf{Diffusion-based Depth and Image Inpainting.} 
After removing the cluttered Gaussian points from the Gaussian scene and obtaining refined masks and rendered RGBD images \( C_{K} = \{c_{j}\}^{K}_{j=1} \) and \( M_{K} = \{m_{j}\}^{K}_{j=1} \) for several keyframes, we proceed with the steps outlined in Infusion~\cite{liu2024infusion}, utilizing a diffusion model and its depth completion model to diffuse the RGBD data of these keyframes. This process yields richly detailed RGB images and smooth, completed depth maps \( C_{K}^{in} = \{c_{j}^{in}\}^{K}_{j=1} \) and \( D_{K}^{in} = \{d_{j}^{in}\}^{K}_{j=1} \), respectively. The challenge lies in leveraging these inconsistent 2D results to supervise the training of the Gaussian scene for a consistent 3D scene. Inspired by~\cite{weder2023removing}, we introduce a mechanism based on the pixel-level uncertainty of primary and secondary viewpoints to harmoniously integrate these inconsistent images, ultimately achieving a complete 3D Gaussian scene \( \Theta_{inpainted} \), thereby overcoming this challenge.

\noindent\textbf{Uncertainty-guided Fine-grained Optimization.} As shown in Fig.~\ref{fig:pipeline}, following the process outlined in Infusion~\cite{liu2024infusion}, we first back-project the RGB image from the primary view into the Gaussian scene using the inpained depth and  images $\{c^{in}_{j}\}^{K}_{j=1}$ and $\{d^{in}_{j}\}^{K}_{j=1}$. However, for large-scale missing regions, relying solely on a single view information for reconstruction is insufficient to ensure reliability across multiple views. Furthermore, the introduction of multiple viewpoints increases inconsistency, as shown in Fig.~\ref{fig:teaser}. To address this challenge, we propose to leverage multi-view uncertainty to assign unfilled regions in the primary view to other views for 3D scene inpainting. Specifically, areas with lower depth values in the primary view are generally associated with higher confidence and clearer details, making them more reliable for reconstruction. Conversely, regions with larger depth values are more difficult to complete and can benefit from complementary information provided by other views. To ensure consistency across regions, we introduce an uncertainty mechanism, initializing the uncertainty within the refined masks on the inpainted depths $D_{K}^{in} = \{d^{in}_{j}\}^{K}_{j=1}$ from key views.

For optimizing each key viewpoint, we adopted a fine-grained optimization strategy guided by uncertainty. Specifically, given the key viewpoint $\Pi_{K}$, the refined masks $M_{K}^{\text{ref}}$ computed in the previous steps, the inpainted images $C_{K}^{in}$, and the predicted depths $D^{in}_{K}$ generated by the diffusion model, we proceed as follows:
For the $j$-th view $\pi_{j} \in \Pi_{K}$,we define fine-grained uncertainty values with resolution $r$ for a key image $c^{in}_{j} \in C^{in}_{K}$, we represent $\mathcal{U}_{j} \in \mathbb{R}^{\frac{H}{r} \times \frac{W}{r}}$, and the uncertainty values are first initialized using the predicted depth $d^{in}_{j} \in D_{K}^{in}$,  as expressed by:
\begin{equation}
\begin{aligned}
\mathcal{U}_{j} [h_{r},w_{r}] = \lambda \cdot \text{mean}(d^{in}_{j}& [h_{r}\times 8:(h_{r}+1)\times 8,\\
& w_{r}\times 8:(w_{r}+1)\times 8 ]), 
\end{aligned}
\end{equation}
where $\lambda$ controls the initialization scale to ensure the optimization process converges within a suitable range, balancing convergence speed and model stability. Here, we perform block-based optimization of the uncertainty values to improve training stability. Point-wise optimization can lead to instability in model optimization.

The confidence weights $\mathcal{W}_{j} \in \mathbb{R}^{H \times W}$ are then defined as:
\begin{equation} 
\begin{aligned}
\mathcal{W}_{j}&[h_{r}\times 8:(h_{r}+1)\times 8, \\
&w_{r}\times 8:(w_{r}+1)\times 8] = \frac{1}{\mathcal{U}_{j}[h_{r},w_{r}]}, 
\end{aligned}
\end{equation}
where $h_{r} \in [0, \frac{H}{r} - 1]$ and $n_{r} \in [0, \frac{W}{r} - 1]$.

The overall loss function is expressed as:
\begin{equation} 
\begin{aligned}
\mathcal{L}_{uncertainty} = \sum_{\pi_{j}\in \Pi_{K}}
[&\left\|m_{j}^{\text{ref}} \odot \frac{\mathcal{W}_{j}^{2}}{2}\odot (c^{r}_{j} - c^{in}_{j})\right\|_{2}^{2} \\
&+ \sum_{m_{j}^{\text{ref}}[h,w]\neq 0} \log \left( \frac{1}{\mathcal{W}_{j}[h,w]}\right) ],
\end{aligned}
\end{equation}
where $M_{j}^{\text{ref}}$ denotes the refined mask and $c^{r}_{j}$ is denoted the rendered image. The optimization focuses exclusively on the uncertainties within the regions defined by the mask. It is worth noting that the second term acts as a regularizer, promoting sparsity in the uncertainty distribution. Ideally, uncertainty should be concentrated in regions far from the viewpoint. A dense uncertainty map would indicate chaotic or unreliable observations, undermining the model’s effectiveness.

With the introduction of the uncertainty loss, our overall loss function is formulated as:
\begin{equation}
\mathcal{L} = \lambda_{1}\mathcal{L}_{rec} + \lambda_{2}\mathcal{L}_{depth} + \lambda_{3}\mathcal{L}_{uncertainty},
\end{equation}
where the coefficients $\lambda_1$, $\lambda_2$, and $\lambda_3$ control the relative contribution of each term. In our experiments, they are empirically set to 1, 0.5, and 1, respectively.

Among them, the reconstruction loss $\mathcal{L}_{rec}$ consists of two components: the reconstruction loss for the background region and the reconstruction loss for the main reference view, which represents the most representative frame among the selected key views. It is defined as:
\begin{equation}
\mathcal{L}_{rec} = \mathcal{L}_{rec}^{bg} + \mathcal{L}_{rec}^{ref}.
\end{equation}

Specifically, the reconstruction constraint for the background region is defined as:
\begin{equation}
\begin{aligned}
\mathcal{L}_{rec}^{bg} = \sum_{\pi_{j} \in \Pi / \Pi_{K}} \Big[ 
& \|\bar{m}_{j}^{r} \odot (c_{j}^{r} - c^{\bar{\Theta}}_{j})\|_{1} \\
& + \text{D-SSIM}(\bar{m}_{j}^{r} \odot c_{j}^{r}, \bar{m}_{j}^{r} \odot c^{\bar{\Theta}}_{j}) \Big].
\end{aligned}
\end{equation}
Here, $\bar{m}_{j}^{r}$ denotes the background region in the refined mask, and $c^{\bar{\Theta}}_{j}$ represents the image rendered from the filtered Gaussians, serving as the supervisory signal.

The reconstruction loss for the main reference view is given by:
\begin{equation}
\begin{aligned}
\mathcal{L}_{rec}^{ref} = 
& \|c_{ref}^{r} - c_{ref}^{in}\|_{1} \\
& + \text{D-SSIM}(c_{ref}^{r}, c_{ref}^{in}) \\
& + \lambda_{4}\text{LPIPS}(c_{ref}^{r}, c_{ref}^{in}).
\end{aligned}
\end{equation}
Here, $c_{\text{ref}}$ denotes the primary view among the selected keyframes and $\lambda_{4}$ is set to 0.5. An additional constraint is applied to this view to enhance reconstruction quality from this critical perspective.

To enforce geometric consistency, we also introduce a depth supervision loss for keyframes:
\begin{equation}
\mathcal{L}_{depth} = \sum_{\pi_{j} \in \Pi_{K}} \|d_{j}^{r} - d_{j}^{in}\|_{1},
\end{equation}
where $d_{j}^{r}$ is the depth map rendered from the Gaussians and $d_{j}^{in}$ denotes the generated pseudo ground-truth depth. Finally, the uncertainty is updated using a gradient descent algorithm, specifically leveraging the Adam optimizer with an initial learning rate of 0.02. Throughout training, inconsistent regions within the key views are progressively refined, leading to updated uncertainty estimates. This dynamic adjustment encourages the model to focus on consistent regions, ultimately balancing multi-view consistency with the preservation of fine-grained details.

Based on the above loss functions, we iteratively update the uncertainty values, as illustrated in Fig.~\ref{fig:uncertainty_map}, which visualizes the final uncertainty distributions across multiple key views. The experimental results indicate that a sparse uncertainty prediction effectively reduces multi-view inconsistency and mitigates the resulting disorder in Gaussian field estimation.

%
%

\label{sec:method}
\begin{figure*} [t]
\centering
\includegraphics[width=0.81\textwidth]{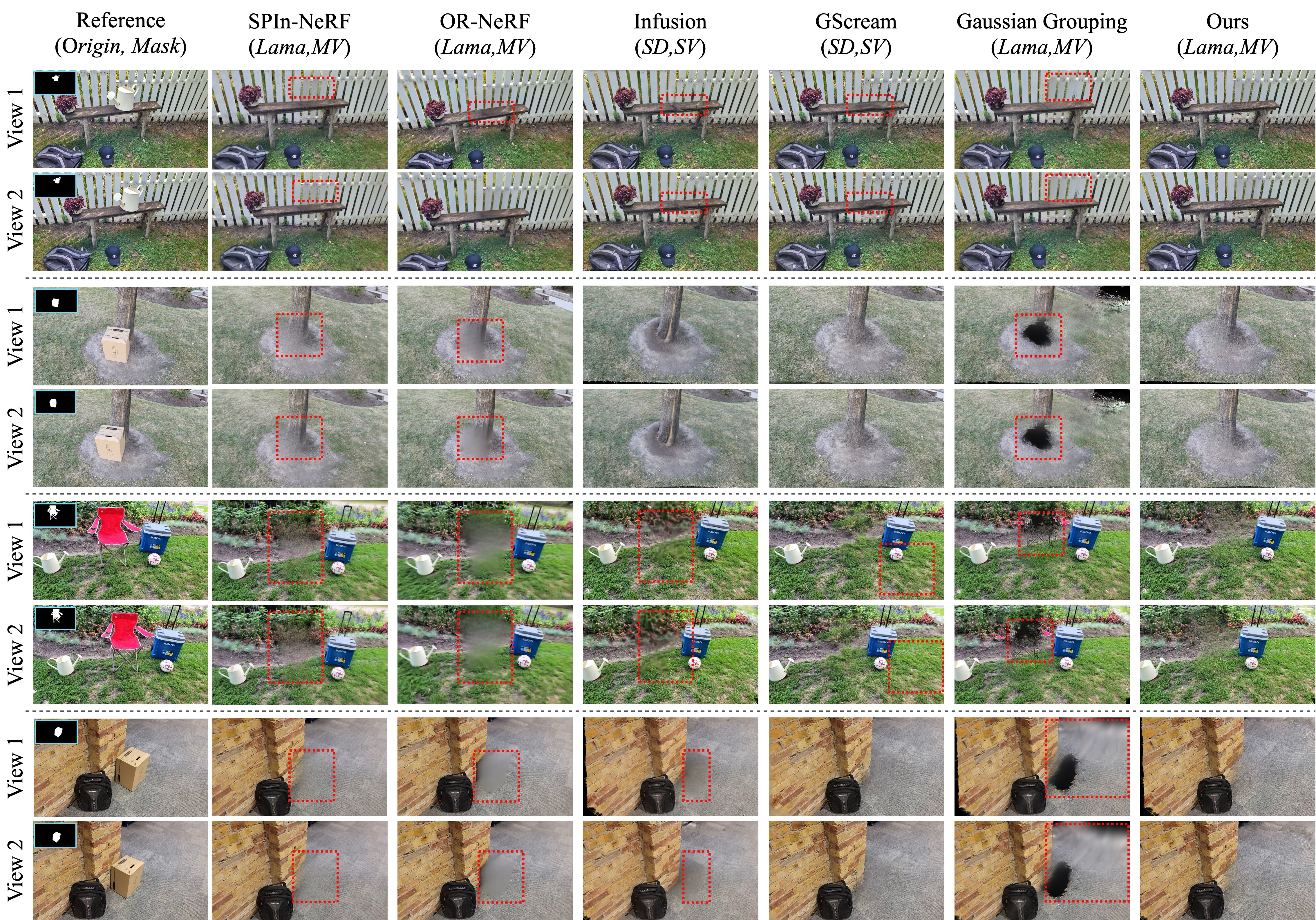}
\caption{Qualitative Comparison of Object Removal and Inpainting Methods. Illustration of rendered images with object removal and inpainting, compared with SPIn-NeRF~\cite{mirzaei2023spin}, OR-NeRF~\cite{yin2023or}, and Gaussian splatting-based methods, including Infusion~\cite{liu2024infusion}, GScream~\cite{wang2024gscream}, and Gaussian Group~\cite{ye2025gaussian}.  Among these methods, all except Infusion~\cite{liu2024infusion} and GScream~\cite{wang2024gscream} use LaMa~\cite{suvorov2022resolution} for image inpainting, while Infusion~\cite{liu2024infusion} and GScream~\cite{wang2024gscream} rely on single-view Stable Diffusion~\cite{rombach2021highresolution} for inpainting. The results highlight the effectiveness of our approach in achieving a natural and seamless object removal effect.}
\label{fig:lama_compare}
\end{figure*}

\section{Experiment}

\subsection{Datasets and Settings}
 To evaluate the effectiveness of our proposed algorithm, we conduct experiments on three representative datasets. Specifically, SPIn-NeRF~\cite{mirzaei2023spin} comprises 10 front-facing wide-field scenes, including 7 outdoor and 3 indoor scenes. Each scene consists of 60 training images and 40 test images, accompanied by binary masks and ground-truth images for object removal evaluation. In addition, we utilize the ``kitchen'' scene from Mip-NeRF360~\cite{barron2022mip} and the ``bear'' and ``garden'' scenes from InNeRF360~\cite{wang2024innerf360}.  These datasets are characterized by large view-angle displacements, covering 360° of camera poses, making them suitable for evaluating the robustness of object removal in challenging scenarios. Due to the lack of ground-truth object-removed images in these two datasets, our evaluation mainly relies on qualitative comparisons to demonstrate the effectiveness of our method under substantial viewpoint changes. Since ground-truth masks are also unavailable in these datasets, we employ SAM-Track~\cite{cheng2023segment} to generate initial rough masks of the target objects for each frame as input for rendering process.

All experiments are conducted on a single RTX 3090 GPU with 24GB VRAM. The initial optimization is performed for 30,000 iterations with a learning rate of 0.02. For the SPIn-NeRF dataset~\cite{mirzaei2023spin}, 4–6 reference views are used during the mask refinement stage, followed by 1,500 iterations of second-stage optimization using 2 sparse views. For larger-scale scenes in other datasets~\cite{barron2022mip,wang2024innerf360}, approximately 10 views are used for refinement, and 4 sparse views are employed for the second stage, which runs for 10,000 iterations.

\subsection{Quantitative Evaluations}
As shown in Tab.~\ref{tab:quantitative}, we present a quantitative comparison of our method against several related approaches. These include NeRF-based methods such as SPIn-NeRF~\cite{mirzaei2023spin} and OR-NeRF~\cite{yin2023or}, as well as Gaussian Splatting-based approaches like Gaussian Grouping~\cite{ye2025gaussian}, GScream~\cite{wang2024gscream}, and Infusion~\cite{liu2024infusion}. Among these methods, only GScream and Infusion utilize single-view Stable Diffusion (SD) for inpainting, thereby avoiding the challenge of multi-view inconsistency. However, this also leads to degraded synthesis quality in distant or novel views, due to the lack of multi-view contextual information and geometric consistency. In contrast, the other methods rely on LaMa~\cite{suvorov2022resolution} to inpaint multiple views, which may introduce noticeable blurriness in the reconstructed scene.

To evaluate the quality of novel view synthesis, we follow previous works and adopt the LPIPS (Learned Perceptual Image Patch Similarity) and FID (Frechet Inception Distance) metrics. Specifically, LPIPS leverages pre-trained deep neural networks to extract image features and computes the perceptual similarity by measuring distances in the feature space, closely aligning with human visual perception. FID, on the other hand, quantifies the statistical difference between feature distributions of real and generated images using a pre-trained Inception network, where lower scores indicate higher visual fidelity.

For a fair comparison, we also integrate LaMa~\cite{suvorov2022resolution} into our pipeline to perform inpainting on sparse views. The experimental results demonstrate that our method achieves competitive performance across most metrics. Although our approach generally outperforms other baselines, the FID score is slightly higher than that of GScream~\cite{wang2024gscream}. This can be attributed to GScream’s use of anchor-based constraints, which help maintain consistency in the Gaussian scene representation and mitigate deviations from the ground-truth distribution. Additionally, our method exhibits clear advantages in computational efficiency, highlighting its practicality for real-world applications.

\subsection{Qualitative Results}
For qualitative evaluation, we compare our method with representative baselines on the SPIn-NeRF dataset. In our setup, two LaMa~\cite{suvorov2022resolution} inpainted views are used for the second-stage optimization. In contrast, most competing methods (except Infusion and GScream) rely on more inpainted views, which may introduce redundancy and multi-view inconsistencies.

SPIn-NeRF~\cite{mirzaei2023spin} and OR-NeRF~\cite{yin2023or}, both NeRF-based methods, suffer from blurry renderings due to limited spatial resolution and weak multi-view consistency. SPIn-NeRF lacks effective cross-view constraints, often leading to appearance artifacts. OR-NeRF removes foregrounds but fails to preserve fine details, with LaMa inpainting yielding overly smoothed results and degraded scene fidelity (see Fig.~\ref{fig:lama_compare}).

Infusion~\cite{liu2024infusion}, relying on single-view depth estimation and back-projection, struggles at mask boundaries, resulting in unrealistic edges and strong artifacts. GScream~\cite{wang2024gscream}, also guided by a single view, often exhibits tearing and ghosting. While it uses anchor point constraints to maintain semantics, these can cause structural duplications and unnatural object extensions (highlighted in red boxes in Fig.~\ref{fig:lama_compare}).

Lastly, Gaussian Grouping~\cite{ye2025gaussian} performs poorly on SPIn-NeRF due to inaccurate object segmentation. The resulting incomplete masks impair inpainting quality, leading to failed reconstructions or heavily blurred outputs.

\begin{table}[htbp]
\centering
\small
\caption{Quantitative results of novel view synthesis after object removal. We conduct a comparative study involving NeRF-based methods (SPIn-NeRF~\cite{mirzaei2023spin} and OR-NeRF~\cite{yin2023or}) and Gaussian Splatting-based approaches (Infusion~\cite{liu2024infusion}, GScream~\cite{wang2024gscream}, and Gaussian Group~\cite{ye2025gaussian}). In this Study, we utilize LaMa~\cite{suvorov2022resolution} for image inpainting to handle missing regions.}
\begin{tabular}{l|c c c}
\toprule
Method & LPIPS$\downarrow$ & FID$\downarrow$ & Time$\downarrow$ \\
\midrule
SPIn-NeRF~\cite{mirzaei2023spin} & 0.47 & 147.31 & 4h\\
OR-NeRF~\cite{yin2023or} & 0.56 & 38.69 &  4h\\
\hline
Gaussian Grouping~\cite{ye2025gaussian} & 0.45 & 123.48 & 20min\\
GScream~\cite{wang2024gscream} & 0.28 & 36.72 & 1.2h\\
Infusion~\cite{liu2024infusion} & 0.42 & 92.62 & 40s\\
Ours &  0.22& 55.17 & 3min\\
\bottomrule
\end{tabular}\label{tab:quantitative}
\end{table}

\begin{figure} [ht]
\centering
\includegraphics[width=0.83\columnwidth]{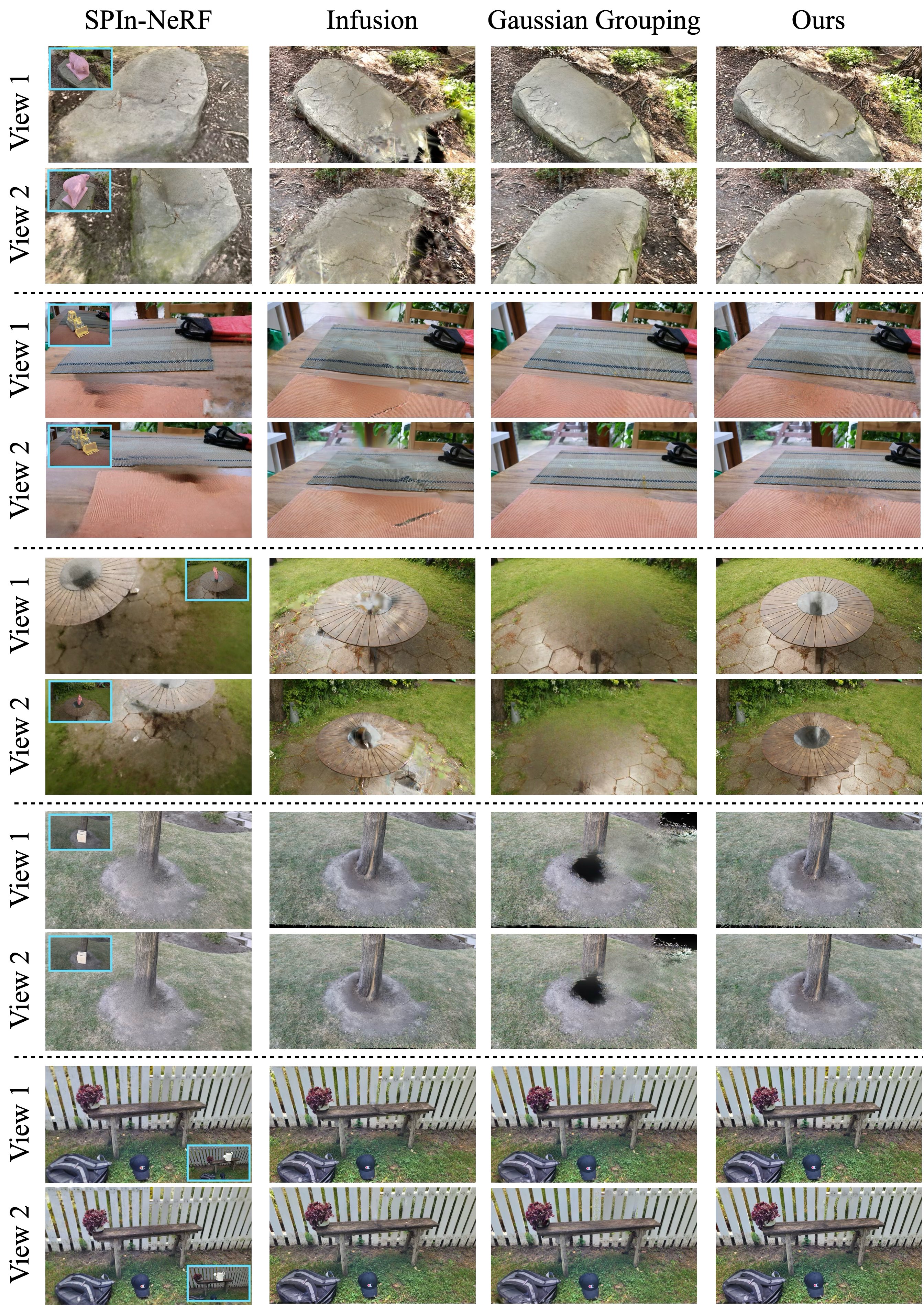}
\caption{Scene Completion Results Using Stable Diffusion-Based Inpainting. For each example, we present the reconstructed scene from two different viewpoints to illustrate the effectiveness of image inpainting in filling missing regions across multiple perspectives. Our method produces consistent and more faithful scene completions, particularly in complex scenarios.}
\label{fig:sd_compare}
\end{figure}

Furthermore, we leverage a diffusion-based inpainting model~\cite{rombach2021highresolution}, to repair occluded regions in the input images. We conduct visual comparisons across several complex scenes to evaluate the effectiveness of our approach. As illustrated in Fig.~\ref{fig:sd_compare}, our method achieves more realistic scene reconstructions with richer texture details. Benefiting from the proposed uncertainty-guided constraint mechanism and more accurate mask estimation, our approach generates multi-view-consistent results that surpass those of Infusion, which relies solely on single-view depth estimation. Notably, our method is able to maintain both high visual clarity and detailed texture continuity across views, demonstrating superior generalization in challenging object removal scenarios.

\begin{table}[ht]
\centering
\caption{Ablation Study of Uncertainty-guided Fine-grained Optimization. Comparison with Non-Uncertainty-guided and Non-Depth Initialized Strategies in Two Scenes.}
\begin{tabular}{@{}lcc@{}}
\toprule
\textbf{Scenes} & LPIPS$\downarrow$ & FID$\downarrow$ \\
\midrule
\multicolumn{3}{l}{\textbf{Scene A}} \\
\quad  w/o uncertainty & 0.22 & 35.58 \\
\quad w/o depth Init. & 0.21 & 31.29 \\
\quad w/ depth + uncertainty & 0.17 & 31.72 \\
\midrule
\multicolumn{3}{l}{\textbf{Scene B}} \\
\quad  w/o uncertainty & 0.21 & 32.63 \\
\quad w/o depth init & 0.21 & 30.28 \\
\quad w/ depth + uncertainty  & 0.18 & 28.04 \\
\bottomrule
\end{tabular}\label{alation:uncertainty_v}
\end{table}

\begin{figure} [ht]
\centering
\includegraphics[width=0.81\columnwidth]{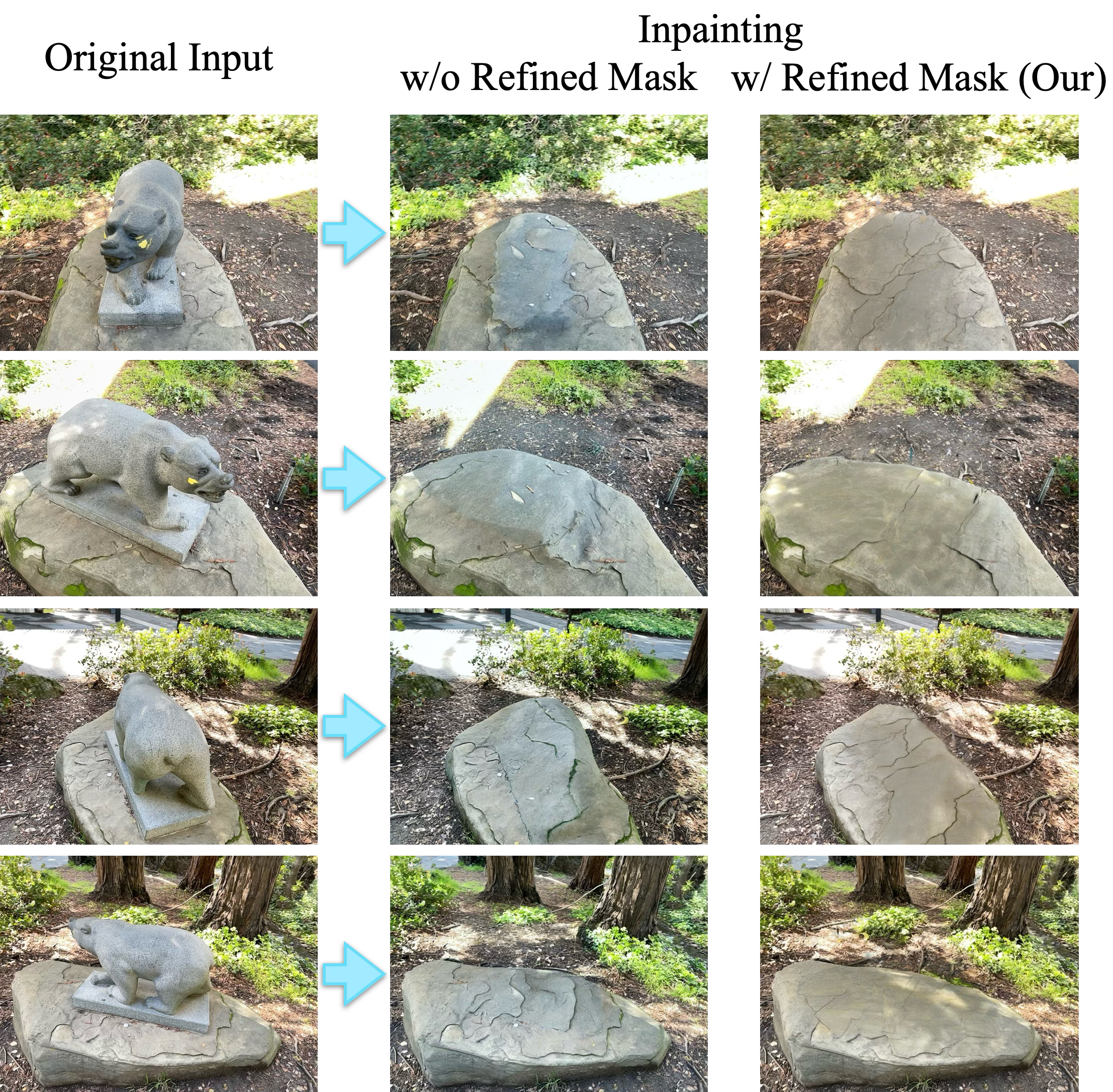}
\caption{Effectiveness of Mask Refinement on Image Inpainting. Qualitative ablation study demonstrating the effectiveness of our mask refinement process for image inpainting. The first column shows the original frames. The second column presents inpainting results using only coarse masks, which often lead to incomplete or unrealistic completions. The third column shows the results with our refined masks, which enable more accurate and visually consistent image restoration, especially in complex scenes.}
\label{fig:ablation_mask_refine}
\end{figure}

\begin{figure} [ht]
\centering
\includegraphics[width=0.83\columnwidth]{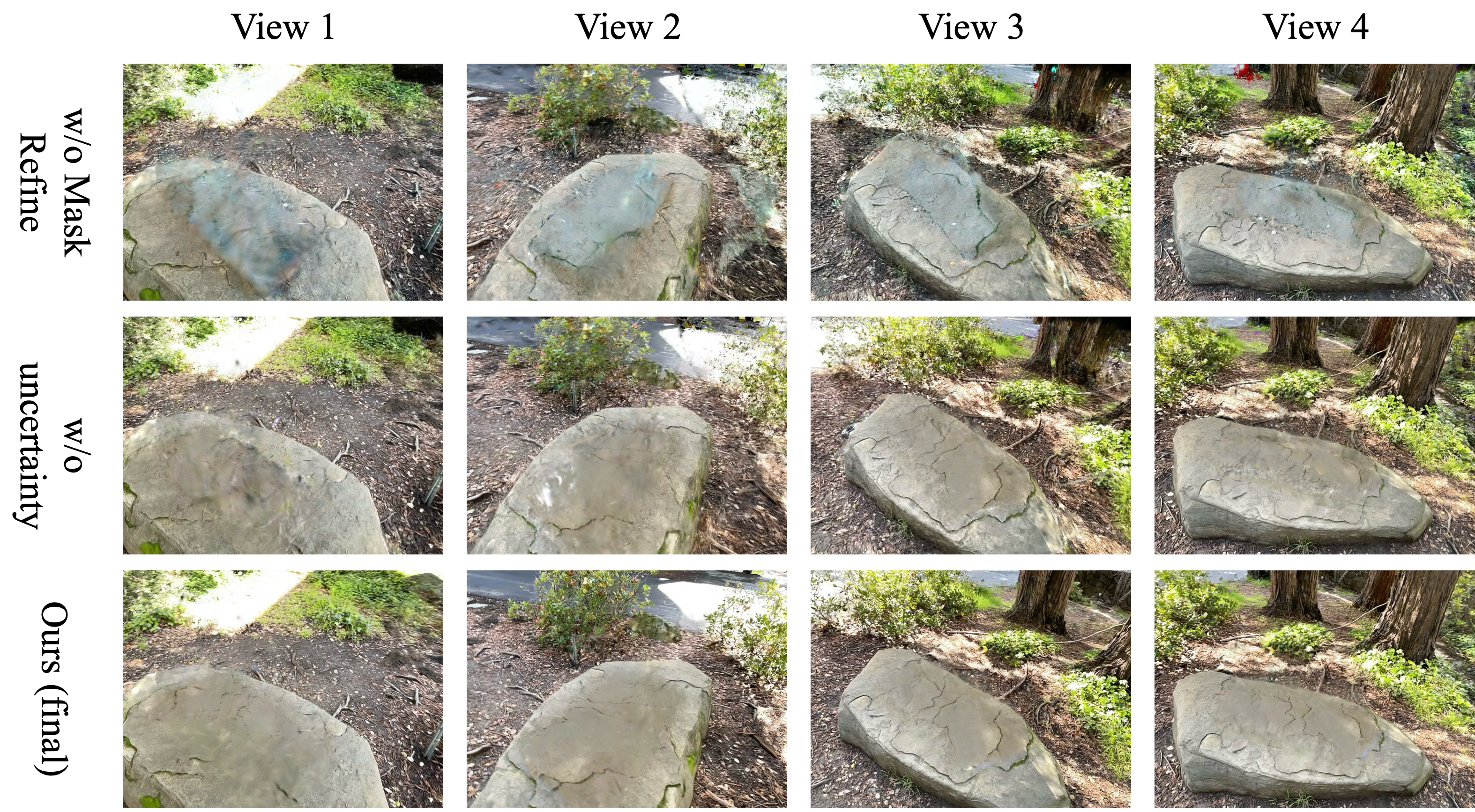}
\caption{Visual Ablation Study on the Effectiveness of Refined Mask and Uncertainty-Guided Optimization in Gaussian Scene Refinement. The first row shows the optimization results without our refined mask strategy, leading to inaccurate and incomplete updates. The second row removes the uncertainty guidance, resulting in suboptimal consistency. The third row displays our full method, demonstrating improved precision and coherence in Gaussian scene reconstruction.}
\label{fig:ablation_rendering_process}
\end{figure}

\begin{figure} [ht]
\centering
\includegraphics[width=0.83\columnwidth]{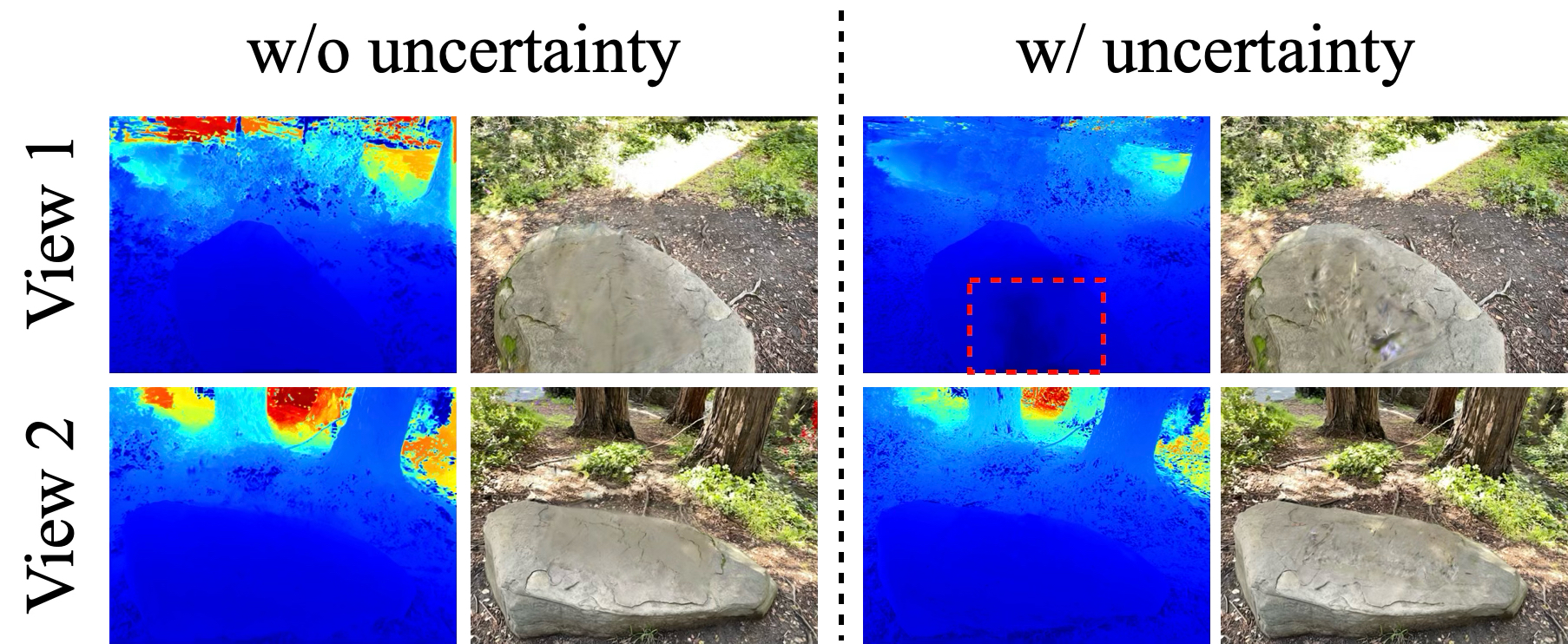}
\caption{Comparison of rendered depth and RGB images from two viewpoints, with and without the proposed uncertainty constraint applied to depth supervision. Results demonstrate that introducing uncertainty into depth supervision destabilizes the optimization process.}
\label{fig:ablation_uncertainty_with_depth}
\end{figure}

\begin{figure*} [ht]
\centering
\includegraphics[width=0.81\textwidth]{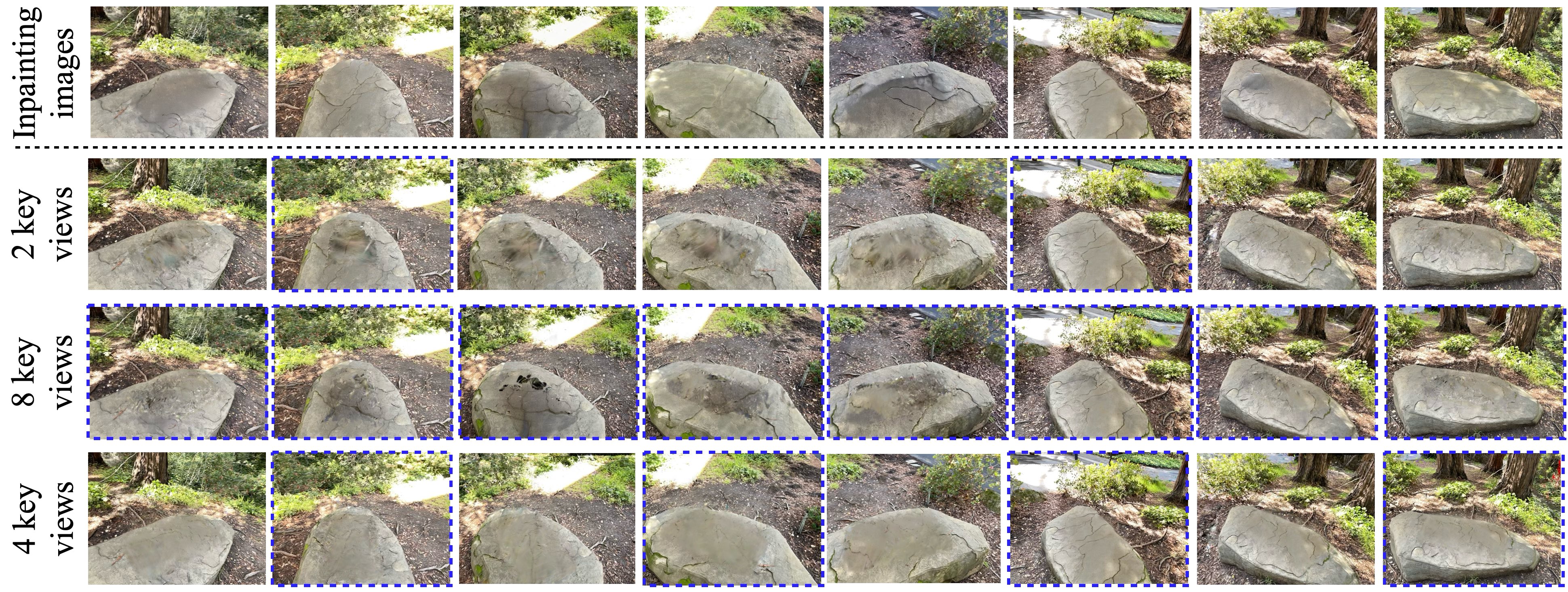}
\caption{Comparison of inpainting results using 2, 4, and 8 key views selected from eight candidate views. The results demonstrate that an appropriate number of key views is critical for inpainting quality, as too few result in incomplete coverage while too many can introduce optimization ambiguity in complex scenes.}
\label{fig:view_selects}
\end{figure*}

\subsection{Ablation Study}
To further validate the effectiveness of our proposed uncertainty-guided optimization strategy, we conducted both quantitative and qualitative ablation studies. Specifically, for the first and last scenes illustrated in Fig.~\ref{fig:lama_compare}, we compare our method against two baseline variants: one without uncertainty guidance and another without using depth estimation for initialization. As shown in Tab.~\ref{alation:uncertainty_v}, our method consistently outperforms both baselines, demonstrating the effectiveness of each design component. In addition, the second row in Fig.~\ref{fig:ablation_rendering_process} provides a qualitative comparison that highlights the role of the uncertainty mechanism. Without uncertainty guidance, training the Gaussian scene using multiple inpainted images leads to inconsistencies in intermediate views. This results in visual artifacts and incoherent textures, particularly in regions farther from the primary views. In contrast, our uncertainty-aware approach enables the network to prioritize supervision from views with high confidence, ensuring that textures from the primary view dominate in shared regions, while complementary views contribute only to filling occluded or missing areas. This strategy helps maintain consistency and realism in the synthesized scene.

To assess the effectiveness of the mask refinement process, we conduct experiments on the ``bear'' scene. As shown in Fig.~\ref{fig:ablation_mask_refine}, utilizing our refined masks during the inpainting stage significantly improves the quality and consistency of the restored content. When using coarse masks directly, the diffusion model tends to overextend into occluded regions behind the removed objects, introducing semantic inconsistencies and hallucinated content. As shown in Fig.~\ref{fig:ablation_mask_refine}, the region behind the sculpture is incorrectly inpainted with semantically unrelated elements, leading to structural conflicts in the final reconstruction. Furthermore, as illustrated in the first row of Fig.~\ref{fig:ablation_rendering_process}, unrefined masks result in inaccurate inpainting, which directly impacts the gaussian scene training by introducing conflicting information between the restored and original regions. This leads to blurred or inconsistent reconstructions. Our mask refinement strategy, on the other hand, better isolates the target object, reduces unnecessary modification of the scene, and thus minimizes artifacts caused by inconsistent supervision.

It is worth noting that the estimated uncertainty is applied only to the RGB images between key views to ensure texture consistency. Depth supervision primarily provides coarse geometric guidance, and minor inconsistencies across views have limited impact on the final reconstruction. In contrast, introducing uncertainty constraints into the depth supervision leads to unstable optimization. As shown in Fig.~\ref{fig:ablation_uncertainty_with_depth}, the rendered depth maps and final Gaussian scenes optimized with uncertainty-guided depth supervision exhibit degraded performance, suggesting that such constraints hinder convergence.

To assess the impact of key view selection on inpainting quality, we conducted visual comparisons using 2, 4, and 8 spatially distributed viewpoints, as illustrated in Fig.~\ref{fig:view_selects}. The results show that using too few views leads to insufficient scene coverage, while using too many can introduce geometric inconsistencies, especially in complex 360-degree environments. Based on these observations, we select approximately $10\%$ of all available views (typically 4 views) as key viewpoints for large-scale scenes, and 2 views for smaller scenes such as those in the SPIn-NeRF dataset. Additionally, about $30\%$ of the views are employed during the mask refinement stage to ensure adequate spatial guidance.

%
%
%

\label{sec:exp}

\section{Limitations and Future Work }
Although the proposed method proves effective, it still has certain limitations. One major challenge is the presence of large and uncontrollable appearance  variations among views. To address this issue, future work may explore video diffusion models guided by auxiliary modalities such as normal maps, depth maps, semantic labels, and texture cues to better preserve geometric and semantic consistency across views. Incorporating a cross-view attention mechanism to diffuse multiple views jointly under consistency constraints, together with Score Distillation Sampling (SDS) loss for auxiliary supervision, could further enhance inter-view coherence. Additionally, extending the framework to 3D object replacement is a promising direction, where joint attention and multi-modal reasoning may ensure consistent and realistic inpainting views across different viewpoints.

\section{Conclusion}
We propose a sparse image-guided 3D Gaussian Inpainting framework. Specifically, we introduce an automatic Mask Refinement module for estimating regions to be inpainted in the initial scene, and to reduce the influence of noisy Gaussian points on the mask estimation, a Gaussian Filtering operation is also applied. This mask refinement technique ensures more accurate mask estimation, significantly improving the boundary quality of the inpainted regions in the scene. Additionally, to address multi-view inconsistencies, we present an Uncertainty-guided Fine-grained Optimization method. This technique automatically estimates the contribution of each pixel to the scene optimization during the Gaussian rendering update process, mitigating conflicts between multi-view images. In our experiments, we demonstrate both quantitatively and qualitatively that our framework can handle scenes from various camera viewpoints and outperforms existing 3D inpainting methods.

\bibliographystyle{IEEEtran}
\bibliography{refs}

\begin{thebibliography}{10}
\providecommand{\url}[1]{#1}
\csname url@samestyle\endcsname
\providecommand{\newblock}{\relax}
\providecommand{\bibinfo}[2]{#2}
\providecommand{\BIBentrySTDinterwordspacing}{\spaceskip=0pt\relax}
\providecommand{\BIBentryALTinterwordstretchfactor}{4}
\providecommand{\BIBentryALTinterwordspacing}{\spaceskip=\fontdimen2\font plus
\BIBentryALTinterwordstretchfactor\fontdimen3\font minus
  \fontdimen4\font\relax}
\providecommand{\BIBforeignlanguage}[2]{{%
\expandafter\ifx\csname l@#1\endcsname\relax
\typeout{** WARNING: IEEEtran.bst: No hyphenation pattern has been}%
\typeout{** loaded for the language `#1'. Using the pattern for}%
\typeout{** the default language instead.}%
\else
\language=\csname l@#1\endcsname
\fi
#2}}
\providecommand{\BIBdecl}{\relax}
\BIBdecl

\bibitem{mildenhall2021nerf}
B.~Mildenhall, P.~P. Srinivasan, M.~Tancik, J.~T. Barron, R.~Ramamoorthi, and
  R.~Ng, ``Nerf: Representing scenes as neural radiance fields for view
  synthesis,'' \emph{Communications of the ACM}, vol.~65, no.~1, pp. 99--106,
  2021.

\bibitem{barron2021mip}
J.~T. Barron, B.~Mildenhall, M.~Tancik, P.~Hedman, R.~Martin-Brualla, and P.~P.
  Srinivasan, ``Mip-nerf: A multiscale representation for anti-aliasing neural
  radiance fields,'' in \emph{Proceedings of the IEEE/CVF international
  conference on computer vision}, 2021, pp. 5855--5864.

\bibitem{wang2021neus}
P.~Wang, L.~Liu, Y.~Liu, C.~Theobalt, T.~Komura, and W.~Wang, ``Neus: Learning
  neural implicit surfaces by volume rendering for multi-view reconstruction,''
  \emph{arXiv preprint arXiv:2106.10689}, 2021.

\bibitem{kerbl20233d}
B.~Kerbl, G.~Kopanas, T.~Leimk{\"u}hler, and G.~Drettakis, ``3d gaussian
  splatting for real-time radiance field rendering.'' \emph{ACM Trans. Graph.},
  vol.~42, no.~4, pp. 139--1, 2023.

\bibitem{huang20242d}
B.~Huang, Z.~Yu, A.~Chen, A.~Geiger, and S.~Gao, ``2d gaussian splatting for
  geometrically accurate radiance fields,'' in \emph{ACM SIGGRAPH 2024
  Conference Papers}, 2024, pp. 1--11.

\bibitem{guedon2024sugar}
A.~Gu{\'e}don and V.~Lepetit, ``Sugar: Surface-aligned gaussian splatting for
  efficient 3d mesh reconstruction and high-quality mesh rendering,'' in
  \emph{Proceedings of the IEEE/CVF Conference on Computer Vision and Pattern
  Recognition}, 2024, pp. 5354--5363.

\bibitem{lu2024scaffold}
T.~Lu, M.~Yu, L.~Xu, Y.~Xiangli, L.~Wang, D.~Lin, and B.~Dai, ``Scaffold-gs:
  Structured 3d gaussians for view-adaptive rendering,'' in \emph{Proceedings
  of the IEEE/CVF Conference on Computer Vision and Pattern Recognition}, 2024,
  pp. 20\,654--20\,664.

\bibitem{chung2023luciddreamer}
J.~Chung, S.~Lee, H.~Nam, J.~Lee, and K.~M. Lee, ``Luciddreamer: Domain-free
  generation of 3d gaussian splatting scenes,'' \emph{arXiv preprint
  arXiv:2311.13384}, 2023.

\bibitem{yi2023gaussiandreamer}
T.~Yi, J.~Fang, J.~Wang, G.~Wu, L.~Xie, X.~Zhang, W.~Liu, Q.~Tian, and X.~Wang,
  ``Gaussiandreamer: Fast generation from text to 3d gaussians by bridging 2d
  and 3d diffusion models,'' in \emph{Proceedings of the IEEE/CVF Conference on
  Computer Vision and Pattern Recognition}, 2024, pp. 6796--6807.

\bibitem{wu20244d}
G.~Wu, T.~Yi, J.~Fang, L.~Xie, X.~Zhang, W.~Wei, W.~Liu, Q.~Tian, and X.~Wang,
  ``4d gaussian splatting for real-time dynamic scene rendering,'' in
  \emph{Proceedings of the IEEE/CVF Conference on Computer Vision and Pattern
  Recognition}, 2024, pp. 20\,310--20\,320.

\bibitem{yang2023real}
Z.~Yang, H.~Yang, Z.~Pan, and L.~Zhang, ``Real-time photorealistic dynamic
  scene representation and rendering with 4d gaussian splatting,'' \emph{arXiv
  preprint arXiv:2310.10642}, 2023.

\bibitem{lin2024gaussian}
Y.~Lin, Z.~Dai, S.~Zhu, and Y.~Yao, ``Gaussian-flow: 4d reconstruction with
  dynamic 3d gaussian particle,'' in \emph{Proceedings of the IEEE/CVF
  Conference on Computer Vision and Pattern Recognition}, 2024, pp.
  21\,136--21\,145.

\bibitem{zhang2025physdreamer}
T.~Zhang, H.-X. Yu, R.~Wu, B.~Y. Feng, C.~Zheng, N.~Snavely, J.~Wu, and W.~T.
  Freeman, ``Physdreamer: Physics-based interaction with 3d objects via video
  generation,'' in \emph{European Conference on Computer Vision}.\hskip 1em
  plus 0.5em minus 0.4em\relax Springer, 2025, pp. 388--406.

\bibitem{xie2024physgaussian}
T.~Xie, Z.~Zong, Y.~Qiu, X.~Li, Y.~Feng, Y.~Yang, and C.~Jiang, ``Physgaussian:
  Physics-integrated 3d gaussians for generative dynamics,'' in
  \emph{Proceedings of the IEEE/CVF Conference on Computer Vision and Pattern
  Recognition}, 2024, pp. 4389--4398.

\bibitem{jiang2024vr}
Y.~Jiang, C.~Yu, T.~Xie, X.~Li, Y.~Feng, H.~Wang, M.~Li, H.~Lau, F.~Gao,
  Y.~Yang \emph{et~al.}, ``Vr-gs: A physical dynamics-aware interactive
  gaussian splatting system in virtual reality,'' in \emph{ACM SIGGRAPH 2024
  Conference Papers}, 2024, pp. 1--1.

\bibitem{jam2021comprehensive}
J.~Jam, C.~Kendrick, K.~Walker, V.~Drouard, J.~G.-S. Hsu, and M.~H. Yap, ``A
  comprehensive review of past and present image inpainting methods,''
  \emph{Computer vision and image understanding}, vol. 203, p. 103147, 2021.

\bibitem{suvorov2022resolution}
R.~Suvorov, E.~Logacheva, A.~Mashikhin, A.~Remizova, A.~Ashukha, A.~Silvestrov,
  N.~Kong, H.~Goka, K.~Park, and V.~Lempitsky, ``Resolution-robust large mask
  inpainting with fourier convolutions,'' in \emph{Proceedings of the IEEE/CVF
  winter conference on applications of computer vision}, 2022, pp. 2149--2159.

\bibitem{rombach2022high}
R.~Rombach, A.~Blattmann, D.~Lorenz, P.~Esser, and B.~Ommer, ``High-resolution
  image synthesis with latent diffusion models,'' in \emph{Proceedings of the
  IEEE/CVF conference on computer vision and pattern recognition}, 2022, pp.
  10\,684--10\,695.

\bibitem{liu2024infusion}
Z.~Liu, H.~Ouyang, Q.~Wang, K.~L. Cheng, J.~Xiao, K.~Zhu, N.~Xue, Y.~Liu,
  Y.~Shen, and Y.~Cao, ``Infusion: Inpainting 3d gaussians via learning depth
  completion from diffusion prior,'' \emph{arXiv preprint arXiv:2404.11613},
  2024.

\bibitem{weder2023removing}
S.~Weder, G.~Garcia-Hernando, A.~Monszpart, M.~Pollefeys, G.~J. Brostow,
  M.~Firman, and S.~Vicente, ``Removing objects from neural radiance fields,''
  in \emph{Proceedings of the IEEE/CVF Conference on Computer Vision and
  Pattern Recognition}, 2023, pp. 16\,528--16\,538.

\bibitem{mirzaei2023spin}
A.~Mirzaei, T.~Aumentado-Armstrong, K.~G. Derpanis, J.~Kelly, M.~A. Brubaker,
  I.~Gilitschenski, and A.~Levinshtein, ``Spin-nerf: Multiview segmentation and
  perceptual inpainting with neural radiance fields,'' in \emph{Proceedings of
  the IEEE/CVF Conference on Computer Vision and Pattern Recognition}, 2023,
  pp. 20\,669--20\,679.

\bibitem{wang2024gaussianeditor}
J.~Wang, J.~Fang, X.~Zhang, L.~Xie, and Q.~Tian, ``Gaussianeditor: Editing 3d
  gaussians delicately with text instructions,'' in \emph{Proceedings of the
  IEEE/CVF Conference on Computer Vision and Pattern Recognition}, 2024, pp.
  20\,902--20\,911.

\bibitem{chen2024gaussianeditor}
Y.~Chen, Z.~Chen, C.~Zhang, F.~Wang, X.~Yang, Y.~Wang, Z.~Cai, L.~Yang, H.~Liu,
  and G.~Lin, ``Gaussianeditor: Swift and controllable 3d editing with gaussian
  splatting,'' in \emph{Proceedings of the IEEE/CVF Conference on Computer
  Vision and Pattern Recognition}, 2024, pp. 21\,476--21\,485.

\bibitem{ye2025gaussian}
M.~Ye, M.~Danelljan, F.~Yu, and L.~Ke, ``Gaussian grouping: Segment and edit
  anything in 3d scenes,'' in \emph{European Conference on Computer
  Vision}.\hskip 1em plus 0.5em minus 0.4em\relax Springer, 2025, pp. 162--179.

\bibitem{huang2024point}
J.~Huang, H.~Yu, J.~Zhang, and H.~Nait-Charif, ``Point'n move: Interactive
  scene object manipulation on gaussian splatting radiance fields,'' \emph{IET
  Image Processing}, 2024.

\bibitem{quan2024deep}
W.~Quan, J.~Chen, Y.~Liu, D.-M. Yan, and P.~Wonka, ``Deep learning-based image
  and video inpainting: A survey,'' \emph{International Journal of Computer
  Vision}, vol. 132, no.~7, pp. 2367--2400, 2024.

\bibitem{elharrouss2020image}
O.~Elharrouss, N.~Almaadeed, S.~Al-Maadeed, and Y.~Akbari, ``Image inpainting:
  A review,'' \emph{Neural Processing Letters}, vol.~51, pp. 2007--2028, 2020.

\bibitem{ballester2001filling}
C.~Ballester, M.~Bertalmio, V.~Caselles, G.~Sapiro, and J.~Verdera,
  ``Filling-in by joint interpolation of vector fields and gray levels,''
  \emph{IEEE transactions on image processing}, vol.~10, no.~8, pp. 1200--1211,
  2001.

\bibitem{tschumperle2005vector}
D.~Tschumperl{\'e} and R.~Deriche, ``Vector-valued image regularization with
  pdes: A common framework for different applications,'' \emph{IEEE
  transactions on pattern analysis and machine intelligence}, vol.~27, no.~4,
  pp. 506--517, 2005.

\bibitem{efros1999texture}
A.~A. Efros and T.~K. Leung, ``Texture synthesis by non-parametric sampling,''
  in \emph{Proceedings of the seventh IEEE international conference on computer
  vision}, vol.~2.\hskip 1em plus 0.5em minus 0.4em\relax IEEE, 1999, pp.
  1033--1038.

\bibitem{barnes2009patchmatch}
C.~Barnes, E.~Shechtman, A.~Finkelstein, and D.~B. Goldman, ``Patchmatch: a
  randomized correspondence algorithm for structural image editing,'' \emph{ACM
  Trans. Graph.}, vol.~28, no.~3, 2009.

\bibitem{darabi2012image}
S.~Darabi, E.~Shechtman, C.~Barnes, D.~B. Goldman, and P.~Sen, ``Image melding:
  Combining inconsistent images using patch-based synthesis,'' \emph{ACM
  Transactions on graphics (TOG)}, vol.~31, no.~4, pp. 1--10, 2012.

\bibitem{huang2014image}
J.-B. Huang, S.~B. Kang, N.~Ahuja, and J.~Kopf, ``Image completion using planar
  structure guidance,'' \emph{ACM Transactions on graphics (TOG)}, vol.~33,
  no.~4, pp. 1--10, 2014.

\bibitem{herling2014high}
J.~Herling and W.~Broll, ``High-quality real-time video inpaintingwith
  pixmix,'' \emph{IEEE Transactions on Visualization and Computer Graphics},
  vol.~20, no.~6, pp. 866--879, 2014.

\bibitem{guo2017patch}
Q.~Guo, S.~Gao, X.~Zhang, Y.~Yin, and C.~Zhang, ``Patch-based image inpainting
  via two-stage low rank approximation,'' \emph{IEEE transactions on
  visualization and computer graphics}, vol.~24, no.~6, pp. 2023--2036, 2017.

\bibitem{wexler2007space}
Y.~Wexler, E.~Shechtman, and M.~Irani, ``Space-time completion of video,''
  \emph{IEEE Transactions on pattern analysis and machine intelligence},
  vol.~29, no.~3, pp. 463--476, 2007.

\bibitem{granados2012background}
M.~Granados, K.~I. Kim, J.~Tompkin, J.~Kautz, and C.~Theobalt, ``Background
  inpainting for videos with dynamic objects and a free-moving camera,'' in
  \emph{Computer Vision--ECCV 2012: 12th European Conference on Computer
  Vision, Florence, Italy, October 7-13, 2012, Proceedings, Part I 12}.\hskip
  1em plus 0.5em minus 0.4em\relax Springer, 2012, pp. 682--695.

\bibitem{newson2014video}
A.~Newson, A.~Almansa, M.~Fradet, Y.~Gousseau, and P.~P{\'e}rez, ``Video
  inpainting of complex scenes,'' \emph{Siam journal on imaging sciences},
  vol.~7, no.~4, pp. 1993--2019, 2014.

\bibitem{huang2016temporally}
J.-B. Huang, S.~B. Kang, N.~Ahuja, and J.~Kopf, ``Temporally coherent
  completion of dynamic video,'' \emph{ACM Transactions on Graphics (ToG)},
  vol.~35, no.~6, pp. 1--11, 2016.

\bibitem{xiangli2022bungeenerf}
Y.~Xiangli, L.~Xu, X.~Pan, N.~Zhao, A.~Rao, C.~Theobalt, B.~Dai, and D.~Lin,
  ``Bungeenerf: Progressive neural radiance field for extreme multi-scale scene
  rendering,'' in \emph{European conference on computer vision}.\hskip 1em plus
  0.5em minus 0.4em\relax Springer, 2022, pp. 106--122.

\bibitem{fridovich2023k}
S.~Fridovich-Keil, G.~Meanti, F.~R. Warburg, B.~Recht, and A.~Kanazawa,
  ``K-planes: Explicit radiance fields in space, time, and appearance,'' in
  \emph{Proceedings of the IEEE/CVF Conference on Computer Vision and Pattern
  Recognition}, 2023, pp. 12\,479--12\,488.

\bibitem{muller2022instant}
T.~M{\"u}ller, A.~Evans, C.~Schied, and A.~Keller, ``Instant neural graphics
  primitives with a multiresolution hash encoding,'' \emph{ACM transactions on
  graphics (TOG)}, vol.~41, no.~4, pp. 1--15, 2022.

\bibitem{Huang2DGS2024}
B.~Huang, Z.~Yu, A.~Chen, A.~Geiger, and S.~Gao, ``2d gaussian splatting for
  geometrically accurate radiance fields,'' in \emph{SIGGRAPH 2024 Conference
  Papers}.\hskip 1em plus 0.5em minus 0.4em\relax Association for Computing
  Machinery, 2024.

\bibitem{guedon2025gaussian}
A.~Gu{\'e}don and V.~Lepetit, ``Gaussian frosting: Editable complex radiance
  fields with real-time rendering,'' in \emph{European Conference on Computer
  Vision}.\hskip 1em plus 0.5em minus 0.4em\relax Springer, 2025, pp. 413--430.

\bibitem{xu2024tiger}
T.~Xu, J.~Chen, P.~Chen, Y.~Zhang, J.~Yu, and W.~Yang, ``Tiger: Text-instructed
  3d gaussian retrieval and coherent editing,'' \emph{arXiv preprint
  arXiv:2405.14455}, 2024.

\bibitem{zhang20243ditscene}
Q.~Zhang, Y.~Xu, C.~Wang, H.-Y. Lee, G.~Wetzstein, B.~Zhou, and C.~Yang,
  ``3ditscene: Editing any scene via language-guided disentangled gaussian
  splatting,'' \emph{arXiv preprint arXiv:2405.18424}, 2024.

\bibitem{liu2022nerf}
H.-K. Liu, I.~Shen, B.-Y. Chen \emph{et~al.}, ``Nerf-in: Free-form nerf
  inpainting with rgb-d priors,'' \emph{arXiv preprint arXiv:2206.04901}, 2022.

\bibitem{mirzaei2023reference}
A.~Mirzaei, T.~Aumentado-Armstrong, M.~A. Brubaker, J.~Kelly, A.~Levinshtein,
  K.~G. Derpanis, and I.~Gilitschenski, ``Reference-guided controllable
  inpainting of neural radiance fields,'' in \emph{Proceedings of the IEEE/CVF
  International Conference on Computer Vision}, 2023, pp. 17\,815--17\,825.

\bibitem{yin2023or}
Y.~Yin, Z.~Fu, F.~Yang, and G.~Lin, ``Or-nerf: Object removing from 3d scenes
  guided by multiview segmentation with neural radiance fields,'' \emph{arXiv
  preprint arXiv:2305.10503}, 2023.

\bibitem{lu2024view}
Y.~Lu, J.~Ma, and Y.~Yin, ``View-consistent object removal in radiance
  fields,'' in \emph{Proceedings of the 32nd ACM International Conference on
  Multimedia}, 2024, pp. 3597--3606.

\bibitem{wang2025learning}
Y.~Wang, Q.~Wu, G.~Zhang, and D.~Xu, ``Learning 3d geometry and feature
  consistent gaussian splatting for object removal,'' in \emph{European
  Conference on Computer Vision}.\hskip 1em plus 0.5em minus 0.4em\relax
  Springer, 2025, pp. 1--17.

\bibitem{huang2023point}
J.~Huang, H.~Yu, J.~Zhang, and H.~Nait-Charif, ``Point'n move: Interactive
  scene object manipulation on gaussian splatting radiance fields,'' \emph{IET
  Image Processing}, 2023.

\bibitem{huang20253d}
S.-Y. Huang, Z.-T. Chou, and Y.-C.~F. Wang, ``3d gaussian inpainting with
  depth-guided cross-view consistency,'' in \emph{Proceedings of the Computer
  Vision and Pattern Recognition Conference}, 2025, pp. 26\,704--26\,713.

\bibitem{wang2024gscream}
Y.~Wang, Q.~Wu, G.~Zhang, and D.~Xu, ``Gscream: Learning 3d geometry and
  feature consistent gaussian splatting for object removal,'' in \emph{ECCV},
  2024.

\bibitem{rombach2021highresolution}
R.~Rombach, A.~Blattmann, D.~Lorenz, P.~Esser, and B.~Ommer, ``High-resolution
  image synthesis with latent diffusion models,'' 2021.

\bibitem{barron2022mip}
J.~T. Barron, B.~Mildenhall, D.~Verbin, P.~P. Srinivasan, and P.~Hedman,
  ``Mip-nerf 360: Unbounded anti-aliased neural radiance fields,'' in
  \emph{Proceedings of the IEEE/CVF conference on computer vision and pattern
  recognition}, 2022, pp. 5470--5479.

\bibitem{wang2024innerf360}
D.~Wang, T.~Zhang, A.~Abboud, and S.~S{\"u}sstrunk, ``Innerf360: Text-guided
  3d-consistent object inpainting on 360-degree neural radiance fields,'' in
  \emph{Proceedings of the IEEE/CVF Conference on Computer Vision and Pattern
  Recognition}, 2024, pp. 12\,677--12\,686.

\bibitem{cheng2023segment}
Y.~Cheng, L.~Li, Y.~Xu, X.~Li, Z.~Yang, W.~Wang, and Y.~Yang, ``Segment and
  track anything,'' \emph{arXiv preprint arXiv:2305.06558}, 2023.

\end{thebibliography}

\end{document}